\documentclass[sn-mathphys,Numbered]{sn-jnl}

\usepackage{subfigure}
\usepackage{graphicx}
\usepackage{multirow}
\usepackage{amsmath,amssymb,amsfonts}
\usepackage{mathtools}
\usepackage{mathrsfs}
\usepackage[title]{appendix}
\usepackage{xcolor}
\usepackage{textcomp}
\usepackage{manyfoot}
\usepackage{booktabs}
\usepackage{algorithm}
\usepackage{algorithmicx}
\usepackage{algpseudocode}
\usepackage{listings}

\usepackage{algorithm}
\usepackage{algpseudocode}
\usepackage{threeparttable}
\usepackage{url}
\usepackage{float}
\usepackage{scalefnt}
\usepackage{lscape}

\usepackage{rotating}
\usepackage{pdflscape}

\usepackage{booktabs, makecell, multirow, tabularx, tabularray}

\usepackage{caption}

\algnewcommand\algorithmicforeach{\textbf{for each}}
\algdef{S}[FOR]{ForEach}[1]{\algorithmicforeach\ #1\ \algorithmicdo}

\begin{document}

\title[Resampling strategies for imbalanced regression: a survey and empirical analysis]{Resampling strategies for imbalanced regression: a survey and empirical analysis}

\author*[1]{\fnm{Juscimara G.} \sur{Avelino}}\email{jga2@cin.ufpe.br}

\author[1]{\fnm{George D. C. } \sur{Cavalcanti}}\email{gdcc@cin.ufpe.br}

\author[2]{\fnm{Rafael M. O.} \sur{Cruz}}\email{rafael.menelau-cruz@etsmtl.ca}

\affil[1]{\orgdiv{Centro de Informática}, \orgname{Universidade Federal de Pernambuco}, \orgaddress{\street{Cidade Universitária}, \city{Recife}, \postcode{50740-560}, \state{PE}, \country{Brazil}}}

\affil[2]{\orgdiv{École de technologie supérieure}, \orgname{Université du Québec}, \orgaddress{\street{1100 Notre Dame St. W.}, \city{Montreal}, \postcode{QC H3C 1K3}, \state{Quebec}, \country{Canada}}}

\abstract{Imbalanced problems can arise in different real-world situations, and to address this, certain strategies in the form of resampling or balancing algorithms are proposed. This issue has largely been studied in the context of classification, and yet, the same problem features in regression tasks, where target values are continuous. This work presents an extensive experimental study comprising various balancing and predictive models, and wich uses metrics to capture important elements for the user and to evaluate the predictive model in an imbalanced regression data context. It also proposes a taxonomy for imbalanced regression approaches based on three crucial criteria: regression model, learning process, and evaluation metrics.
The study offers new insights into the use of such strategies, highlighting the advantages they bring to each model's learning process, and indicating directions for further studies. The code, data and further information related to the experiments performed herein can be found on GitHub: \url{https://github.com/JusciAvelino/imbalancedRegression}.}


\maketitle

\section{Introduction}

\begin{figure}[h]
    \centering
    \includegraphics[width=10cm]{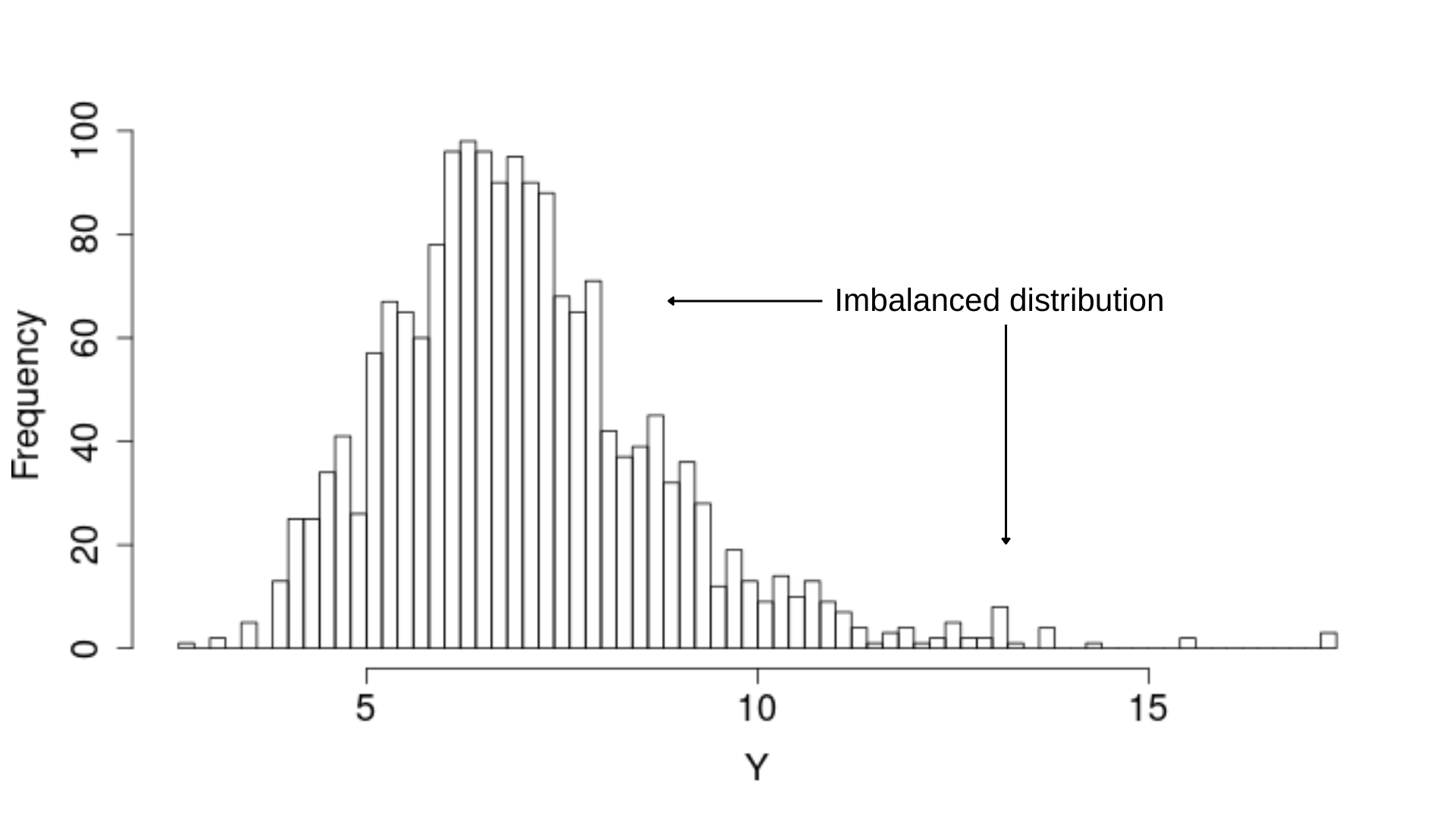}
    \caption{Distribution and frequency of the target value Y from the FuelCons dataset.}
    \label{fig:imb}
\end{figure}

Imbalanced datasets are often encountered in multiple real-world applications. For classification tasks, such an issue has been studied~\cite{haixiang2017learning, krawczyk2016learning, johnson2019survey}. Nonetheless, it is also present in regression tasks~\cite{branco2016ubl}. Branco, Torgo, and Ribeiro~\cite{branco2017smogn} define imbalanced problems based on the simultaneity of two factors: i)~a disproportionate preference of the user at the domain of the target variable, and ii)~insufficient representation of the data available in the most relevant cases for the user. In classification tasks, an imbalanced dataset is determined through the presence of a class having a smaller representation (minority class) than another one (majority class). However, in regression problems, the target value is continuous, thus representing a complex definition, because the target value is not constrained to a limited set of discrete values, unlike in classification problems where the target value represents specific categories or classes. Figure~\ref{fig:imb} presents the distribution and frequency of examples drawn from an imbalanced dataset (FuelCons) with target values ranging from 2.7 to 17.3. To analyze this range, we employed a bin width of approximately 0.2, resulting in a total of 74 bins. The values at the chart's edges show little frequency and are considered rare examples. In this context, Ribeiro~\cite{ribeiro2011utility} proposes the concept of a relevance function which determines the relevance of continuous target values in defining certain examples as rare and others as normal. This definition allows to verify an imbalanced between instances considered rare and those seen as normal.

Standard regression tasks assume that all values of the domain are of equal importance, and are typically evaluated based on the performance of the most frequent values. However, values that are little represented are often extremely relevant, not only to the user, but also in the prediction process. For example, in the context of software engineering prediction mistakes in large projects are associated with higher development costs~\cite{rathore2017linear}, whereas during temperatures prediction in a meteorological application, errors that surface while predicting extreme conditions (e.g., very high temperatures) are even much more costly~\cite{ribeiro2020imbalanced}.
This scenario presents particular difficulties for learning algorithms, which tend to follow the interval of values in greater quantity while neglecting the rare ones in the distribution. Hence, failing to obtain a good prediction performance for these particular examples. 

Studies looking at solutions for imbalanced regression problems have faced relatively little scrutiny when compared to those related to classification problems~\cite{haixiang2017learning}. The most common approach used to address this gap has been to modify the distribution of examples by balancing the training data before the actual learning process begins. Some of these strategies are Random Under-sampling~\cite{torgo2013smote}, which removes examples from intervals having greater quantities, Random Over-sampling~\cite{branco2019pre}, which replicates rare values in the dataset, and the WEighted Relevance-based Combination Strategy (WERCS)~\cite{branco2019pre}, which creates a weighted combination biased versions of the under- and over-sampling strategies. In addition, several real-world imbalanced regression problems rely on resampling strategies to properly deal with rare and extreme cases, such as in software defect prediction~(\cite{bal2018cross},~\cite{bal2020wr},~\cite{rathore2017linear} and~\cite{rathore2017towards}) and Enzyme Optimum Temperature prediction \cite{gado2020improving}, as well as to assist in detecting arsenic concentration in soil using satellite imagery\cite{agrawal2021detecting}. Hence, the variety of problems and increased interest in this field demonstrates the need for studies on imbalanced regression techniques.


Another difficulty encountered in such scenarios is related to the fact that traditional performance metrics, such as the Mean Squared Error (MSE) and the Mean Absolute Error (MAE), do not adequately capture user-defined criteria~\cite{branco2019pre}. Additionally, recent works have proposed new performance metrics for evaluating the performance of regression models under imbalanced target distributions, and place greater emphasis on errors occurring in rare cases. In these cases, Precision, Recall, and F1-score metrics, as described for regression tasks~\cite{torgo2009precision}, and the squared error-relevance area (SERA) metric proposed in~\cite{ribeiro2020imbalanced}, are commonly used. Nevertheless, a comparison between multiple imbalanced regression strategies under these performance metrics, and of how they differ in their approach to assessing the model's performance, is still an open question.

Therefore, our main goal is to analyze the effects of resampling strategies for dealing with imbalanced regression problems from different perspectives. To this end, we conduct an extensive experimental study employing different resampling strategies and learning algorithms. In addition, we use metrics that can assess the models' performance in imbalanced regression tasks, such as the F1-score for regression and SERA~\cite{ribeiro2020imbalanced}. To the best of our knowledge, this is the first work that performs a comprehensive empirical analysis of resampling techniques for imbalanced regression tasks. In contrast, for imbalanced classification tasks, numerous surveys and empirical studies have evaluated resampling algorithms in different scenarios, such as binary problems~\cite{garcia2020understanding, kovacs2019empirical, wojciechowski2017difficulty, roy2018study, ali2019review, del2015analysis, diez2015diversity, moniz2021no}, multiclass classification~\cite{cruz2019dynamic, saez2016analyzing}, and data streams~\cite{aguiar2022survey, zyblewski2019data}.

The broad scope of our experimental analysis, which considers multiple resampling strategies, regression models, and performance metrics, is at the core of the uniqueness of our research since it allowed us to assess the relationship among these three variables. Our study thus differs from \cite{branco2016survey}, which addresses only theoretical aspects of imbalanced problems in general. Moreover, regarding the performance metrics, using the SERA metric~\cite{ribeiro2020imbalanced} is highlighted since no other work has evaluated all resampling strategies using it specifically. 
 
The following research questions guide this study: i)~Is it worth using resampling strategies? ii) Which resampling strategies influence predictive performance the most? iii) Does the choice of best strategy depend on the problem, the learning model, and the metrics used? iv) Does the number of training examples resulting from each strategy influence the results? v) Do the features of the data (percentage of rare cases, number of rare cases, dataset size, number of attribues and imbalance ratio) impact the predictive performance of the models? The experimental analysis revealed that resampling strategies are beneficial to the vast majority of regression models. The best strategies include GN, RO, and WERCS. Another important point is that choosing the best strategy depends on the dataset, the regression model, and the metric used when evaluating the system's performance. Furthermore, we found that the dataset size, the number of rare cases, the number of attribute and the imbalance ratio significantly influence the results. The smallest datasets and those with the fewest rare cases are the most challenging. Models demonstrate superior performance in datasets with fewer features. Lastly, concerning the imbalance ratio, regression models encounter more significant challenges with a higher imbalance ratio.

\subsubsection*{Contributions}

\begin{itemize}

  \item We propose a novel taxonomy for imbalanced regression tasks according to the regression model, learning strategy and metrics.
  
  \item We review the main strategies used for imbalanced regression tasks.
  
  \item We conduct an extensive experimental study comparing the performance of state-of-the-art resampling strategies and their effects on multiple learning algorithms and novel performance metrics proposed in the literature.
  
  \item We analyze the impact of dataset characteristics (e.g., dataset size and the number of rare cases) on the model's predictive performance.

\end{itemize}

This work is organized as follows: Section~\ref{sec:concepts} presents the basic concepts and proposes a taxonomy for imbalanced regression problems. Section~\ref{sec:resampling} describes the resampling approaches evaluated in this study highlighting their advantages and disadvantages. Section~\ref{sec:experiments} presents the experimental methodology by describing the data, algorithms, parameters, and performance metrics used in this work. Results are shown in Section~\ref{sec:results}. Section~\ref{sec:lessons} presents the lessons learned by revisiting and answering the research questions. Finally, Section~\ref{sec:conclusion} brings our conclusions.

\section{Basic Concepts and Proposed Taxonomy}
\label{sec:concepts}

Some fundamental concepts must be grasped in order to understand the notion of imbalanced regression. In this context, the concept of relevance function is presented herein and a taxonomy is proposed to organize the strategies required. The relevance function is a fundamental concept in imbalanced regression, as it defines the importance of each sample in the dataset. Finally, a taxonomy is proposed to categorize the approaches used to address imbalanced regression problems, providing a way to understand the existing literature. Based on this taxonomy, we review the main strategies for dealing with imbalanced regression problems.

\subsection{Relevance Function}
\label{sec:relevance}

The concept of relevance function is crucial when it comes to understanding the imbalanced regression problem and some strategies for dealing with it. Proposed by Ribeiro~\cite{ribeiro2011utility}, the relevance function ($\phi : Y \rightarrow [0,1]$) determines the relevance of the examples in each dataset using an automatic method. The relevance value determines the examples that are normal and those that are rare, with the rare ones being the least represented in the dataset. The intuition of the relevance function is to automatically set the significance of data points within a dataset by assigning relevance scores. In this way, the relevance function serves as the foundation for evaluating models in the context of imbalanced regression, as well as for data resampling. Consequently, using a different relevance function alters both the model evaluation and data resampling.

To the best of our knowledge, this definition of relevance function is unique in the literature. In \cite{ribeiro2011utility} and \cite{ribeiro2020imbalanced}, the relevance function is showcased using the Piecewise Cubic Hermite Interpolating Polynomials (pchip) and cubic spline methods. However, it was noted that cubic spline interpolation cannot provide precise control over the function. It fails to confine the relevance function within the specified [0, 1] interval scale. This limitation is rectified by the pchip method, employing suitable derivatives at control points, thereby ensuring properties like positivity, monotonicity, and convexity. Consequently, \cite{ribeiro2011utility} proposed relevance function utilizes the pchip method and aligned with this, the works in the field utilize this function.


The relevance function ($\phi$) is calculated using Piecewise Cubic Hermite Interpolating Polynomials (pchip)~\cite{dougherty1989nonnegativity} over a set of control points (Algorithm~\ref{alg:pchip}). The algorithm receives as input the control points~($S$) with their respective relevance values~($\varphi(y_k)$) and derivative~($\varphi'(y_k)$). The condition $y1 < y2 < ... < ys$ ensures that the data points are ordered in ascending order of their y-values. This ordering is fundamental for properly functioning the pchip algorithm. As a result, the algorithm produces a separate $\phi(y)$ polynomial for each interval $[y_k, y_{k+1}]$, with coefficients calculated based on the control points and their derivatives within that specific interval, where the variable $k$ represents the index for the input set $S$ control points.

\begin{algorithm}[H]
     \caption{pchip($S$): Piecewise Cubic Hermite Interpolating Polynomials}
     \label{alg:pchip}
     
    \flushleft \hspace{0.3cm} \textbf{Input:} 
     
    \hspace{0.6cm} $S=\{\langle y_k, \varphi(y_k), \varphi'(y_k)\rangle\}^{s}_{k=1}$, with $y_1<y_2<...<y_s$, relevance values $\varphi(y_k)$ and preliminary derivative $\varphi'(y_k)$ \\
    
    \hspace{0.3cm} \textbf{Output:} 
    
    \hspace{0.6cm} $\phi(y)$: a Piecewise Cubic Hermite Interpolating Polynomial

    \begin{algorithmic}[1]

    \For{$k \gets 1$ \textbf{to} $s-1$} 
    \State $h_k \gets y_{k+1} - y_k$
    \State $\delta_k \gets (\varphi(y_{k+1}) - \varphi(y_k))/h_k$
    \State $a_k \gets \varphi(y_k)$
    \EndFor
    \State $\{b_k\}^{s-1}_{k=1} \gets$ \textit{check\_slopes} $(\{\varphi'(y_k)\}^{s-1}_{k=1}, \{\delta_k\}^{s-1}_{k=1})$ \Comment{Monotone Cubic Spline}
    \For{$k \gets 1$ \textbf{to} $s-1$}
    \State $c_k \gets (3\delta_k - 2b_k + b_{k+1})/h_k$
    \State $d_k \gets (b_k - 2\delta_k + b_{k+1})/h^2_k$
    \EndFor
    \State \Return $\phi(y) = a_k +b_k (y - y_k) + c_k (y-y_k)^2 + d_k (y-y_k)^3, y \in [y_k,y_{k+1}]$
    
\end{algorithmic}
\end{algorithm}

The control points can be defined based on domain knowledge or provided by an automated method. When control points are defined based on domain knowledge, selecting them is guided by the expertise and understanding of the specific problem or dataset. This approach relies on the insights and experience of individuals familiar with the data and its context. Ideally, access to domain knowledge for defining control points would be preferred. However, this knowledge is often unavailable or nonexistent \cite{ribeiro2020imbalanced}. Therefore, the utilization of an automatic method for control point definition becomes necessary. An example of defining control points of the NO2 emissions problem based on domain knowledge is presented in Table~\ref{tab:NO2}. Control points are determined based on Directive 2008/50/EC. The objective is to maintain the $LNO2$ (target) hourly concentration values below a limit equal to $ln(150 \mu g/m^3) \approx 5.0$, indicating maximum relevance, and the annual average guideline of $ln(40 \mu g/m^3) \approx 3.7$, indicating minimal relevance. And the lowest $LNO2$ concentration value $ln(3 \mu g/m^3) \approx 1.1$ is attributed minimal relevance.

\begin{table}[h]
\caption{Control points of LNO2 concentration thresholds
according to Directive 2008/50/EC \cite{ribeiro2020imbalanced}.}
\begin{tabular}{@{}llll@{}}
\toprule
$y_k$ : $LNO2$ concentration values &                     & $\phi(y_k)$ & $\phi'(y_k)$ \\ \midrule
Low concentration:                  & $ln(3 \mu g/m^3) \approx 1.1$  & 0.0         & 0.0          \\
Annual mean guideline:              & $ln(40 \mu g/m^3) \approx 3.7$  & 0.0         & 0.0          \\
Limit threshold:                    & $ln(150 \mu g/m^3) \approx 5.0$ & 1.0         & 0.0          \\ \bottomrule
\end{tabular}
\label{tab:NO2}
\end{table}

In this work, we employ the automatic method, proposed by \cite{ribeiro2011utility}, to define the control points. This method is based on Tukey's boxplot \cite{tukey1970exploratory}. The Tukey's boxplot is a graphical representation used to display the distribution of a dataset through its five summary statistics: The adjacent limits $adj_L$ (Eq. \ref{adjL}) and $adj_H$ (Eq. \ref{adjH}), first quartile ($Q1$), third quartile ($Q3$) and median $\Tilde{Y}$ (Eq. \ref{Y}). In turn, the control points are defined by the adjacent limits and the median value. The input to the pchip algorithm consists of control points, their relevance and derivatives. For this purpose, to the adjacent values ($adj_L$, $adj_H$) maximum relevance is assigned, which equals 1, and the median value ($\Tilde{Y}$) with relevance value equal to zero. All control points are initialized with derivative $\phi'(y_k)$ equal to 0. In addition to defining the control points using Tukey's boxplot, Ribeiro et al.~\cite{ribeiro2020imbalanced} proposes the utilization of the adjusted boxplot, as proposed by Hubert et al.~\cite{hubert2008adjusted}.

\begin{equation}
\label{adjL}
    adj_L = Q1 - 1.5 \cdot IQR
\end{equation}

\begin{equation}
\label{adjH}
    adj_H = Q3 + 1.5 \cdot IQR
\end{equation}

\begin{equation}
\label{Y}
    \Tilde{Y} = \mbox{median of } Y
\end{equation}

\noindent where $Q1$ and $Q3$ are the first and third quartile, respectively, and $IQR = Q3 - Q1$.

\begin{figure} [h]
    \centering
    \includegraphics[width=10cm]{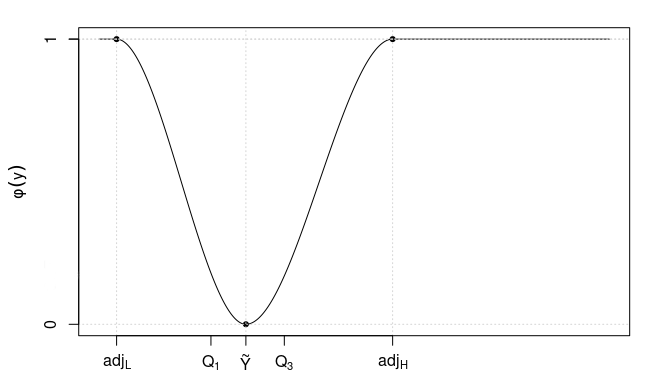}
    \caption{Relevance function of the fuelCons dataset.}
    \label{fig:relevanceFunction}
\end{figure}

Figure~\ref{fig:relevanceFunction} illustrates the relevance function resulting from the pchip algorithm, for the fuelCons dataset. The points approaching $\Tilde{Y}$ have negligible relevance, whereas points that move away from $\Tilde{Y}$ and approach $adj_L$ or $adj_H$ have maximum relevance.

\begin{algorithm}[!ht]
    \caption{check\_slopes ($\Phi, \Delta$) \cite{fritsch1980monotone}}
    \label{alg:check_slopes}
    
    \flushleft \hspace{0.3cm} \textbf{Input:} 
    
    \hspace{0.6cm} $\Phi$ = $\{\varphi'(y_k)\}^{s-1}_{k=1}$, $\Delta = \{\delta_k\}^{s-1}_{k=1}$ \\
    
    \flushleft \hspace{0.3cm} \textbf{Output:} 
    
    \hspace{0.6cm} $\Phi$: Modified derivative values \\
    
    \begin{algorithmic}[1]
        
        \For{$k \gets 1$ \textbf{to} $s-1$}
            \If{$\delta_k = 0$} \Comment{Initialize the derivatives}
                \State $\varphi'(y_k) \gets \varphi'(y_{k+1}) \gets 0$
            \Else
                \State $\alpha \gets \varphi'(y_k)/\delta_k$
                \State $\beta\gets \varphi'(y_{k+1})/\delta_k$
                \If{$\varphi'(y_k) \neq 0 \wedge \alpha < 0 $} \Comment{select an underset to preserve monotonicity}
                    \State $\varphi'(y_k) \gets -\varphi'(y_k)$
                    \State $\alpha \gets \varphi'(y_k)/\delta_k$
                \EndIf
                \If{$\varphi'(y_{k+1}) \neq 0 \wedge \beta < 0 $} \Comment{select an underset to preserve monotonicity}
                    \State $\varphi'(y_{k+1}) \gets -\varphi'(y_{k+1})$
                    \State $\beta \gets \varphi'(y_{k+1})/\delta_k$
                \EndIf
                \State{$\tau_1 \gets 2\alpha + \beta - 3$}
                \State{$\tau_2 \gets \alpha + 2\beta - 3$}
                \If{$\tau_1 > 0 \wedge \tau_2 > 0 \wedge \alpha(\tau_1+\tau_2) < \tau_1\tau_2$}  \Comment{modifying the derivative values}
                    \State$\tau \gets 3\delta_k/\sqrt{\alpha^2+\beta^2}$
                    \State$\varphi'(y_k) \gets \alpha\tau$
                    \State$\varphi'(y_{k+1}) \gets \beta\tau$
                \EndIf
            \EndIf
        \EndFor
        \State \Return $\Phi = \{\varphi'(y_k)\}^{s-1}_{k=1}$
        
    \end{algorithmic}
\end{algorithm}

The interpolation generates a function that crosses the control points. One of the main goals is to learn the correct slopes in the data points such that the interpolant is monotonic by parts. To this end, a method that implements the Monotone Cubic Spline~\cite{fritsch1980monotone} (line 6) is used. The check\_slopes method~(Algorithm~\ref{alg:check_slopes}) ensures that the derivative is zero when the control point for a maximum or minimum local~\cite{ribeiro2020imbalanced}.

A relevance threshold ($t_R$) defined by the user is employed to divide the data into rare ($D_R$) and normal ($D_N$) values. Given a dataset $D$, the sets $D_R$ and $D_N$ are defined considering the superior and inferior thresholds as follows: $D_R = \{\langle \textbf{x},y \rangle \in D: \phi(y) \geq t_R\}$ and $D_N = \{\langle \textbf{x},y \rangle \in D:\phi(y) < t_R\}$.

\subsection{Proposed Taxonomy}
\label{sec:taxonomy}

In the context of class imbalanced problems, solutions are often classified into four groups: Algorithmic level, Cost-sensitive, Ensemble learning, and Data preprocessing~\cite{galar2011review, lopez2013insight}. However, one problem with this classification is that there is a significant overlap between the ensemble learning, data preprocessing, and cost-sensitive groups. Ensemble learning approaches can be used in conjunction with any other approaches by learning the base models, accounting to target imbalance at the algorithmic level, or applying data preprocessing prior to training each base model in the ensemble. Therefore, to better understand the different approaches for dealing with imbalanced regression problems, we can categorize the strategies into three main groups: i)~Regression Models, ii)~Learning Process Modification, and iii)~Evaluation Metrics.


\begin{figure}[h]
    \centering
    \includegraphics[width=11cm]{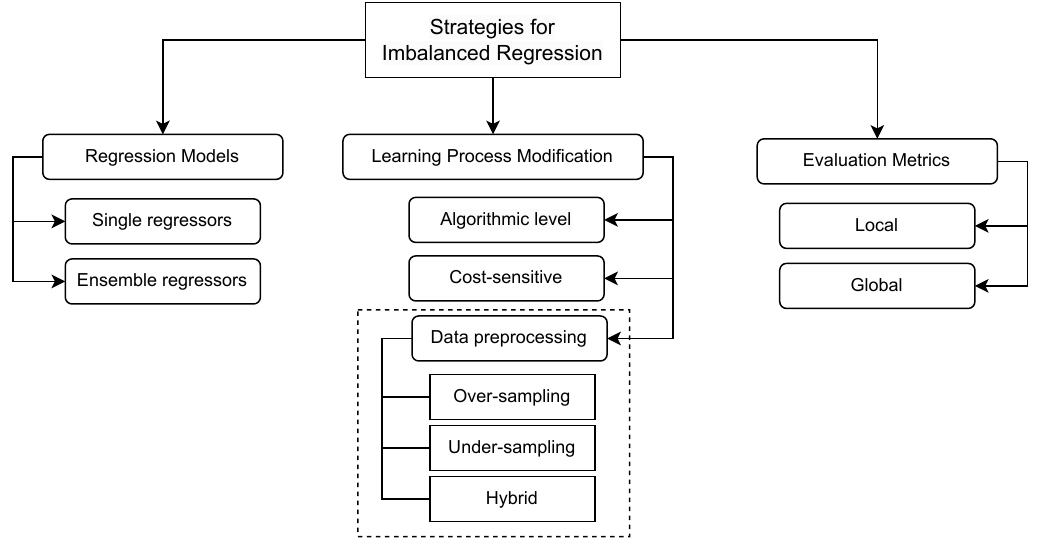}
    \caption{Proposed taxonomy for imbalanced regression problems.}
    \label{fig:taxonomy}
\end{figure}

The first group of strategies comprises regression models, such as single models and ensembles, which can be used to address imbalanced regression problems. However, their performance can be further improved by incorporating data preprocessing, cost-sensitive learning, and algorithmic-level modifications. The second group describes these additional strategies which can help adjust the learning process to deal with the target imbalance, thus leading to better results when compared to using the models alone. The third group  comprises the evaluation  metrics and is divided into local and global subgroups. The local metrics require a relevance threshold to distinguish extreme values and conduct a local evaluation, and thus, cases with a relevance score lower than the threshold are disregarded. Conversely, global metrics do not require a relevance threshold, making a global evaluation, considering all the examples. To conclude, categorizing these strategies into three groups can provide a better understanding of the approaches and enable the selection of the most suitable strategy for dealing with imbalanced regression problems. As shown in Figure~\ref{fig:taxonomy}, data preprocessing takes the spotlight, which is the main focus of this work. Herein, we explore and compare different data preprocessing techniques to improve the performance of regression models (single models and ensembles) in imbalanced regression problems.

\subsubsection*{Regression Models}

Regression models such as MLPRegressor, Linear Support Vector Regression (SVR) and decision trees can be used to solve problems with imbalanced regression data, but they may not perform well due to the imbalance. In such cases, it may be necessary to utilize other techniques such as data preprocessing or cost-sensitive learning, or to modify the algorithm, to address the issue. In the same perspective, ensemble models, such as bagging, boosting, and random forest, can also be utilized in addressing these problems. Solutions based on ensemble learning combined with data preprocessing strategies and cost-sensitive were proposed. In~\cite{branco2018rebagg} the REsampled BAGGing (REBAGG) model was proposed in a bid to integrate data resampling strategies with bagging, and had the advantage of generating a diverse set of models taking into account the different ways training data are resampled using the Random Under-sampling, Random Over-sampling and SmoteR strategies. SMOTEBoost~\cite{moniz2018smoteboost} includes a resampling step when boosting, where SmoteR is used to direct the distribution of data towards rare cases. In the same context, \cite{moniz2017evaluation} carried out a performance study of ensemble methods in regression tasks with imbalanced datasets.

\subsubsection*{Learning Process Modification}

Learning Process Modification refers to the techniques used to modify the training process of machine learning algorithms to take into account rare cases. These techniques include algorithmic level modification, as well as the cost-sensitive and data preprocessing methods. At an algorithmic level, a model is introduced in~\cite{torgo2003predicting} with new division criteria for the regression trees that allow to induce trees at extreme and rare predicted values. Yang et al.~(2021)~\cite{yang2021delving} proposed methods aimed at favoring the similarity between near targets by applying a kernel distribution to soften the distribution in the target and space of attributes.~\cite{ribeiro2011utility} then addressed a utility-based algorithm involving cost-sensitive learning designed with a set of rules extracted from the generation of different regression trees aimed at obtaining accurate and interpretable predictions for imbalanced regression. \cite{steininger2021density} proposed a density-based weighting approach to address the issue of imbalanced regression, building on the cost-sensitive method. This approach assigns higher weights to rare cases by taking into account their local densities. Finally, one of the most common approaches for treating imbalanced issues is data preprocessing, also known as resampling or balancing algorithms, which precede the learning process, altering the examples distribution. The method works by either removing samples from common cases (i.e., under-sampling) or generating synthetic samples for rare events (i.e., oversampling). Data processing techniques have the advantage of allowing the use of just about any learning algorithm concurrently, without affecting the explicability of the model~\cite{branco2019pre}. 

Different resampling strategies have been proposed to deal with imbalanced regression problems. Most such techniques are based on existing resampling strategies proposed for classification problems. That is the case, for example, of the SmoteR algorithm, which is a variation of the Smote~\cite{chawla2002smote} algorithm, with the following main adaptations made to adjust to the issue of regression: i)~the definition of rare cases, ii)~the creation of synthetic examples, and iii)~the definition of target values for newly generated examples. Also on the basis of the Smote algorithm, \cite{camacho2022geometric} proposed Geometric SMOTE, which generates synthetic data points along the line connecting two existing data points. Other strategies adapted from imbalanced classification are: Random Under-sampling~\cite{torgo2013smote}, based on the idea of \cite{kubat1997addressing}; Random Over-sampling~\cite{branco2019pre}, proposed for the classification in \cite{batista2004study}, and Introduction of Gaussian Noise~\cite{branco2019pre}, adapted from \cite{lee1999regularization, lee2000noisy}. In contrast, the SMOGN (SmoteR with Gaussian Noise)~\cite{branco2017smogn} and the WERCS (WEighted Relevance-based Combination Strategy)~\cite{branco2019pre} strategies were originally proposed for handling imbalanced regression problems. Furthermore, \cite{song2022distsmogn} introduced a distributed version of the SMOGN called DistSMOGN. The method uses a weighted sampling technique to generate synthetic samples for rare cases, in addition to considering the data distribution in each node of the distributed system. For the imbalanced data streams in regression models context, \cite{aminian2021chebyshev} introduced two sampling strategies (ChebyUS, ChebyOS) based on the Chebyshev inequality to improve the performance of existing regression methods on imbalanced data streams. The approaches use a weighted learning strategy that assigns higher weights to rare cases in order to balance the training process.

Each strategy resamples data differently. However, they appear to be based on the same principles: reducing normal examples and/or increasing rare examples. Under-sampling, which reduces normal examples, is the basis of the Random Under-sampling strategy. In contrast, over-sampling, which increases rare examples, can have a simple performance, as in Random Over-sampling, or by generating synthetic cases, as in the SmoteR Algorithm and Introduction of Gaussian Noise. Other strategies are based on the aforementioned models. Examples include the SmoteR with Gaussian Noise (SMOGN), which combines the Random Under-sampling strategy with the SmoteR and Introduction of Gaussian Noise over-sampling strategies. Also, the WEighted Relevance-based Combination Strategy (WERCS) combines the Random Under-sampling  and  Random Over-sampling strategies by using weights to perform the resampling without establishing a relevance threshold.

In our study, we analyze a variety of data preprocessing techniques to optimize the performance of single and ensemble regression models in addressing imbalanced regression problems. Our objective is to compare the effectiveness of different approaches in identifying the most suitable strategies for this situation. By carefully assessing these techniques, we aim to provide guidance as to how to increase the success rate of regression models using data preprocessing techniques in imbalanced regression tasks.

\subsubsection*{Evaluation Metrics}

The choice of assessment metrics is fundamental in an imbalanced datasets scenario. Some metrics, such as the MSE, may fool users when the focus is on the accuracy of rare values of the target variable~\cite{moniz2014resampling} since it does not consider the relevance of each testing example. To show the limitations of the MSE metric and how the scores obtained by different metrics can significantly differ, we present a synthetic example (Table~\ref{tab:ilustmetrics}). For 10 examples in the \textit{FuelCons} dataset, we present hypothetical predictions for two artificial models: $M_1$ and $M_2$. The True row represents the true target for each instance in the dataset, directly obtained from the FuelCons dataset. The $\phi$ row is the relevance value of each example. Meanwhile, the $M_1$ and $M_2$ rows showcase predictions generated by the respective models for individual test examples. In parallel, the $M_1$ and $M_2$ loss rows quantify the differences between the true target and the predictions made by the models for each test example. The example shows that $M_1$ generates more accurate predictions for the less relevant examples, which are less represented in the dataset, while $M_2$ performs better for more relevant examples, which are more frequently represented. Nonetheless, if the models' performances are assessed using the MSE metric, there will be no difference in scores between them. This is because the MSE metric considers all examples as having the same relevance ($\phi$). Therefore, for the imbalanced data scenario, where each example has a particular relevance, it is more interesting to use metrics that consider the relevance of each particular example.

\begin{table}[h]
\caption{Predictions of two artificial models}
\begin{tabular}{|ccccccccccc|}
\hline
\multicolumn{1}{|l}{\textbf{}}   & \multicolumn{10}{c|}{\textbf{Test examples}}  \\
\hline
\multicolumn{1}{|c|}{True}       & \multicolumn{1}{c|}{2.70}          & \multicolumn{1}{c|}{3.20}          & \multicolumn{1}{c|}{3.50}          & \multicolumn{1}{c|}{4.10}          & \multicolumn{1}{c|}{4.50}          & \multicolumn{1}{c|}{4.70}          & \multicolumn{1}{c|}{5.20}          & \multicolumn{1}{c|}{5.70}          & \multicolumn{1}{c|}{9.20}          & 17.30         \\
\multicolumn{1}{|c|}{$\phi$}     & \multicolumn{1}{c|}{\textbf{0.00}} & \multicolumn{1}{c|}{\textbf{0.00}} & \multicolumn{1}{c|}{\textbf{0.00}} & \multicolumn{1}{c|}{\textbf{0.00}} & \multicolumn{1}{c|}{\textbf{0.00}} & \multicolumn{1}{c|}{\textbf{0.02}} & \multicolumn{1}{c|}{\textbf{0.57}} & \multicolumn{1}{c|}{\textbf{1.00}} & \multicolumn{1}{c|}{\textbf{1.00}} & \textbf{1.00} \\ \hline
\multicolumn{1}{|c|}{$M_1$}      & \multicolumn{1}{c|}{2.66}          & \multicolumn{1}{c|}{3.14}          & \multicolumn{1}{c|}{3.40}          & \multicolumn{1}{c|}{3.80}          & \multicolumn{1}{c|}{4.00}          & \multicolumn{1}{c|}{3.80}          & \multicolumn{1}{c|}{4.10}          & \multicolumn{1}{c|}{4.40}          & \multicolumn{1}{c|}{7.70}          & 15.50         \\
\multicolumn{1}{|c|}{$M_1$ Loss} & \multicolumn{1}{c|}{\textbf{0.04}} & \multicolumn{1}{c|}{\textbf{0.06}} & \multicolumn{1}{c|}{\textbf{0.10}} & \multicolumn{1}{c|}{\textbf{0.30}} & \multicolumn{1}{c|}{\textbf{0.50}} & \multicolumn{1}{c|}{\textbf{0.90}} & \multicolumn{1}{c|}{\textbf{1.10}} & \multicolumn{1}{c|}{\textbf{1.30}} & \multicolumn{1}{c|}{\textbf{1.50}} & \textbf{1.80} \\ \hline
\multicolumn{1}{|c|}{$M_2$}      & \multicolumn{1}{c|}{0.90}          & \multicolumn{1}{c|}{1.70}          & \multicolumn{1}{c|}{2.20}          & \multicolumn{1}{c|}{3.00}          & \multicolumn{1}{c|}{3.60}          & \multicolumn{1}{c|}{4.20}          & \multicolumn{1}{c|}{4.90}          & \multicolumn{1}{c|}{5.60}          & \multicolumn{1}{c|}{9.14}          & 17.26         \\
\multicolumn{1}{|c|}{$M_2$ Loss} & \multicolumn{1}{c|}{\textbf{1.80}} & \multicolumn{1}{c|}{\textbf{1.50}} & \multicolumn{1}{c|}{\textbf{1.30}} & \multicolumn{1}{c|}{\textbf{1.10}} & \multicolumn{1}{c|}{\textbf{0.90}} & \multicolumn{1}{c|}{\textbf{0.50}} & \multicolumn{1}{c|}{\textbf{0.30}} & \multicolumn{1}{c|}{\textbf{0.10}} & \multicolumn{1}{c|}{\textbf{0.06}} & \textbf{0.04} \\ \hline
\end{tabular}
\begin{tablenotes}

\item True - Target values
\item $\phi$ - Relevance values
\item M1 and M2 - Model predictions
\item M1 Loss and M2 Loss - Prediction errors

\end{tablenotes}
\label{tab:ilustmetrics}
\end{table}

Other metrics consider each example as having a particular relevance score, such as Precision, Recall, and the F1-score, which were proposed for regression applications in~\cite{torgo2009precision}. In addition, the Squared error‐relevance area (SERA) metric, which was specifically created for imbalanced regression, was proposed by Ribeiro et al.~\cite{ribeiro2020imbalanced}. This metric aims to effectively assess the model's performance for predictions of extreme values while being robust to model bias. Table~\ref{tab:estimated}  presents the MSE, F1-score, and SERA values for the example presented in Table~\ref{tab:ilustmetrics}. As earlier mentioned, for the MSE, the models are regarded as equals since they both have the same error amplitude. Nonetheless, for the F1-score and SERA, which consider each example's relevance, $M_2$ is the best model as it presents a lower error in the most important examples.

\begin{table}[!h]
\caption{Performances of two artificial models}
\begin{tabular}{@{}cccc@{}}
\toprule
\multicolumn{4}{c}{\textbf{Estimated Performance}}                \\ \midrule
\textbf{}      & \textbf{MSE} & \textbf{F1-score} & \textbf{SERA} \\
\textbf{$M_1$} & 0.955        & 0.598             & 7.885         \\
\textbf{$M_2$} & 0.955        & 0.983             & 0.076         \\ \bottomrule
\end{tabular}
\label{tab:estimated}
\end{table}

The Precision, Recall, and F1-score metrics require that a relevance threshold be defined to determine extreme values. Thus, a local evaluation is performed, since examples below the threshold are ignored. Furthermore, these metrics use the concept of a utility-based framework~\cite{torgo2007utility},~\cite{ribeiro2011utility}. Such a structure uses the numeric error of the prediction and the relevance of the actual and predicted values. The utility of predicting a value $\hat{y}$ for $y$ is calculated from the notions of costs and benefits of numeric predictions~\cite{branco2019pre}, and thus, the utility function $U^p_\phi(\hat{y},y)$ is given by Equation~\ref{Up}, where $\hat{y}$ is the predicted value and $y$ is the actual value.

\begin{equation}  
\label{Up}
  \begin{multlined}  
   U^p_\phi(\hat{y},y) = B_\phi(\hat{y},y) - C^p_\phi(\hat{y},y) \\\  
      = \phi(y)\cdot(1-\Gamma_B(\hat{y},y))-\phi^p(\hat{y},y)\cdot\Gamma_C(\hat{y},y)  
  \end{multlined} 
\end{equation}

The utility is given by the difference between the prediction benefit ($B_\phi(\hat{y},y)$) and cost ($C^p_\phi(\hat{y},y)$) of prediction $\hat{y}$ for $y$. The benefit is defined as a proportion of the relevance of the actual value according to the following equation: $\phi(y)\cdot(1-\Gamma_B(\hat{y},y))$, where $\Gamma_B(\hat{y},y)$ is the bounded loss function (Equation~\ref{loss}). This equation defines a loss function, $\Gamma_B(\hat{y},y)$, which quantifies the loss incurred when making a prediction $\hat{y}$ for the actual value $y$ (Equation \ref{AD}). This loss function operates on a scale from 0 to 1, where 0 represents no loss, and 1 represents maximum loss.

\begin{equation}
\label{loss}
    \Gamma_B(\hat{y},y)) = 
    \begin{cases}
        L(\hat{y},y)/\dot{L}_B(\hat{y},y), & \mbox{if } L(\hat{y},y) < \dot{L}_B(\hat{y},y)\\
        1, & \mbox{if } L(\hat{y},y) \ge \dot{L}_B(\hat{y},y)
    \end{cases}
\end{equation}

\noindent $L$ is a ``standard" loss function (e.g., absolute deviation (Equation~\ref{AD})) and $\dot{L}_B$ is the benefit threshold function, (Equation~\ref{limitbenefit}). The benefit threshold function identifies the point at which the predicted value ceases to provide a benefit.  This can happen because of two conditions: (i) surpassing the maximum acceptable loss of the bump or (ii) being situated on a different bump \cite{ribeiro2011utility}.

\begin{equation}
\label{AD}
    L(\hat{y}, y) = |\hat{y} - y|
\end{equation}

\begin{equation}
\label{limitbenefit}
    \dot{L}_B(\hat{y}, y) = min\{ b^\Delta_{\gamma(y)}, \ddot{L}_B(\hat{y}, y) \}
\end{equation}

\noindent where $b^\Delta_{\gamma(y)}$ is the maximum admissible loss, defined in Equation~\ref{delta}. The maximum admissible loss is calculated for each bump $i$. A bump refers to a interval of the domain, denoted as $B \subseteq Y$ \cite{ribeiro2011utility}. $b^-$ is the mean value at which the target variable reaches the minimum relevance before reaching its maximum value, and $b^*$ is the mean value at which the target variable reaches the maximum relevance. The reason for this definition is that this function is contingent upon the smallest discrepancy concerning the target variable when transitioning from the most pertinent value within a bump ($b^*_i$) to an alternative bump. The smallest differences regarding the target variable can have two effects on model performance. On the positive side, it can make the model more accurate by focusing on the areas where predictions must be very close to the actual values. This is useful when you need high accuracy in specific parts of the data. Conversely, the model might become too fixated on the training data, making it sensitive to unusual data points and not very good at handling new data, leading to overfitting. Consequently, this implies that when dealing with ``narrow" bumps, our sensitivity to prediction errors is heightened, whereas for broader bumps, we are more inclined to deem larger disparities between the actual and forecasted values as acceptable \cite{ribeiro2011utility}. 

\begin{equation}
\label{delta}
b^\Delta_{\gamma(y)} = 2 \cdot \mbox{min}\{\mid b^-_i - b^*_i\mid, \mid b^*_i - b^-_{i+1} \mid \}
\end{equation}

\begin{figure}[h]
        \centering
        \includegraphics[width=10cm]{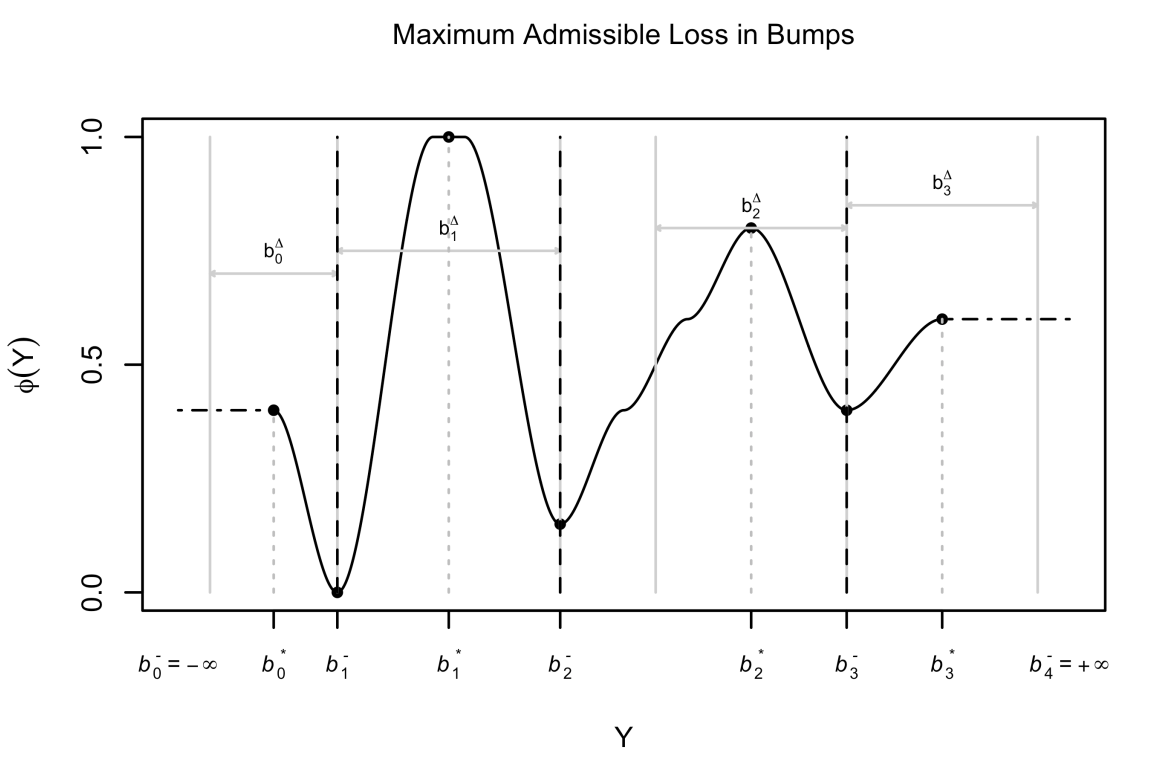}
        \caption{Bumps partition of $Y$ with respect to relevance function $\phi$ and the maximum admissible loss in bumps. Each bump $i$ is characterized by its partition node $b^-$ and by one global maximum $b^*$. Each bump has a maximum error tolerance defined by the double of the smalles amplitude in the bump between each of one of its bounds and its maximum value \cite{ribeiro2011utility}.}
        \label{fig:bump}
    \end{figure}

Figure~\ref{fig:bump} shows the bump partition obtained for a relevance function and the maximum admissible loss for each bump. This arbitrary relevance function, defined in the context of non-uniform utility regression, has four quite diﬀerent bumps.

And $\ddot{L}_B(\hat{y},y))$ (Equation~\ref{eq9}) is defined as follows:

\begin{equation}
\label{eq9}
    \ddot{L}_B(\hat{y},y)) = 
    \begin{cases}
        \mid y - b^-_{\gamma(y)}\mid, & \mbox{if } \hat{y}< y)\\
        \mid y - b^-_{\gamma(y)+1}\mid, & \mbox{if } \hat{y} \ge y)
    \end{cases}
\end{equation}

This definition satisfies two essential conditions: (1) The initial component within the min function addresses the maximum allowable error range within the true value's context, guaranteeing a level of reasonable accuracy in the prediction; (2) The subsequent component within the min function evaluates whether the predicted value aligns with the correct action by considering its proximity to the boundaries of the context associated with the true value.

The cost is given by the mean of weighted relevance ($\phi^p(\hat{y},y)$) (Equation~\ref{op}), where the parameter $p$ is used to define the weights between the two relevances and $\Gamma_C(\hat{y},y)$ is the bounded loss function in the scale [0;1]. This equation calculates the weighted relevance of the predicted value $\hat{y}$ and the actual value $y$. The parameter $p$ defines the weights between these two relevances. The intuition here is to balance the predicted value's importance and the utility function's actual value.

\begin{equation}
\label{op}
    \phi^p(\hat{y},y) = (1-p)\phi(\hat{y})+p\phi(y)
\end{equation}

The cost function $\Gamma_C(\hat{y},y)$ is calculated according to Equation~\ref{perdac}.

\begin{equation}
\label{perdac}
    \Gamma_C(\hat{y},y)) = 
    \begin{cases}
        L(\hat{y},y)/\dot{L}_C(\hat{y},y), & \mbox{if } L(\hat{y},y) < \dot{L}_C(\hat{y},y)\\
        1, & \mbox{if } L(\hat{y},y) \ge \dot{L}_C(\hat{y},y)
    \end{cases}
\end{equation}

where $L$ is the standard loss function, and $\dot{L}_C$ is the cost threshold function (Equation~\ref{limitcost}): 

\begin{equation}
\label{limitcost}
    \dot{L}_C(\hat{y}, y) = min\{ b^\Delta_{\gamma(y)}, \ddot{L}_C(\hat{y}, y) \}
\end{equation}

and $\ddot{L}_C(\hat{y},y))$ is defined as follows:

\begin{equation}
    \ddot{L}_C(\hat{y},y)) = 
    \begin{cases}
        \mid y - b^*_{\gamma(y)-1}\mid, & \mbox{if } \hat{y}< y)\\
        \mid y - b^*_{\gamma(y)+1}\mid, & \mbox{if } \hat{y} \ge y)
    \end{cases}
\end{equation}

Captured using the utility function, the Precision and Recall metrics are defined by Equations~\ref{precision} and~\ref{recall}, respectively.

\begin{equation}
\label{precision}
    Precision = \frac{\sum_{\phi(\hat{y}_i)>t_R} (1+U^p_\phi(\hat{y}_i, y_i))}{\sum_{\phi(\hat{y}_i)>t_R}(1+\phi(\hat{y}_i))}
\end{equation}

\begin{equation}
\label{recall}
    Recall = \frac{\sum_{\phi(y_i)>t_R} (1+U^p_\phi(\hat{y}_i, y_i))}{\sum_{\phi(y_i)>t_R}(1+\phi(y_i))}
\end{equation}

The relevance of the actual value $y_i$ is defined by $\phi(y_i)$, as defined in Section \ref{sec:relevance}, and $\phi(\hat{y}_i)$ is the relevance of the predicted value $\hat{y}_i$.  $t_R$ is a threshold defined by the user for the relevance values, and $U^p_\phi(\hat{y}_i, y_i)$ is the utility function previously described.

The Precision and Recall metrics can be aggregated in compound measures, such as F1-score, defined by Equation~\ref{f1}:t

\begin{equation}
\label{f1}
     \textit{F1-score} = \frac{(\beta^2+1) \cdot Precision \cdot Recall}{\beta^2 \cdot Precision + Recall}
\end{equation}

\noindent where $0 \leq \beta \leq 1$ controls the relative importance of the Recall for the Precision. These compound measures have the advantage of allowing comparisons between models by providing a single score~\cite{torgo2009precision}.

These metrics require the definition of an ad-hoc relevance threshold and do not consider examples below the threshold for model evaluation \cite{ribeiro2020imbalanced}. To address this, \cite{ribeiro2020imbalanced} proposed the SERA metric.

SERA metric can assess models' efficacy and optimize them for predicting rare and extreme cases. This metric does not require a definition of a relevance threshold and thus performs a global evaluation since all data points are considered. The Squared error‐relevance is obtained in relation to a cutting $t$ achieved based on a relevance function $\phi : Y \rightarrow [0,1]$. A subset $D^t = \{ \langle \textbf{x},y \rangle \in D: \phi(y) \ge t\}$ formed based on the cutting $t$ is considered for this estimate, such as in Equation~\ref{ser}:

\begin{equation}
\label{ser}
    SER_t = \sum\limits_{i \in D^t}(\hat{y_i}-y_i)^2
\end{equation}

The Squared error‐relevance area (SERA) represents the area below the curve $SER_t$, obtained through integration presented in Equation~\ref{sera}: 

\begin{equation}
\label{sera}
    SERA = \int\limits_{0}^{1} SER_t \hspace{1mm} dt = \int\limits_{0}^{1} \sum\limits_{i\in D^t}(\hat{y_i}-y_i)^2 \hspace{1mm} dt
\end{equation}

The $SER_t$ curve offers a broad view of prediction errors in the domain at various relevance cutoff values. Therefore, a smaller area under the curve (SERA) indicates a better model. It is noteworthy that assuming uniform preferences with $\phi(y) = 1$, SERA is comparable with the sum of squared errors.

\section{Resampling Strategies}
\label{sec:resampling}

The most common way to deal with imbalanced datasets is to use resampling strategies changing the data distribution to balance the targets~\cite{moniz2017resampling}. Such strategies are concentrated on the following three main approaches: i)~over-sampling, ii)~under-sampling, and iii)~a combination of these two approaches. In over-sampling, rare cases are generated to compensate for the imbalanced distribution. The Random Over-sampling technique~\cite{branco2019pre} is an example of such a technique, which works by replicating rare cases prior to training. However, it is also possible to perform over-sampling by generating synthetic cases, as in the SmoteR~\cite{torgo2013smote} and Introduction of Gaussian Noise strategies~\cite{branco2019pre}.

Conversely, under-sampling techniques aim to exclude larger quantity data(i.e., normal examples). The Random Under-sampling algorithm~\cite{torgo2013smote} uses this notion. Some strategies employ a combination of approaches, such as the SmoteR and Introduction of Gaussian Noise, which generates synthetic cases and uses under-sampling, WEighted Relevance based Combination Strategy~\cite{branco2019pre} , thus combining the approaches of under-sampling and over-sampling. The SMOGN~\cite{branco2017smogn} uses the generation of synthetic cases with SmoteR and GN and under-sampling.

Sections~\ref{sec:SmoteR}-\ref{sec:WERCS} provide an overview of the resampling strategies evaluated in this work. These strategies were selected based on their wide adoption in the literature. Conversely, other strategies were disregarded due to an absence of publicly available source code for them, limited reproducibility, and infrequent utilization by researchers for diverse problem domains. Finally, Section~\ref{sec:advantages} critically analyzes the resampling strategies with a visual example.

\subsection{SmoteR}
\label{sec:SmoteR}

The SMOTE for regression (SmoteR) algorithm was proposed in~\cite{torgo2013smote} (Algorithm~\ref{alg:ST}). Like the other methods addressing imbalanced regression issues, it requires a relevance function ($\phi(y)$) and a relevance threshold ($t_R$). The relevant or unimportant examples are defined from such a function. The algorithm removes the least relevant examples (lines 4 to 7), which are considered ``normal", and then generates synthetic examples based on the most relevant examples (line 8). The generation process basically follows the idea in the SMOTE, namely, first selecting one rare case from the dataset as the seed case and one of its K-Nearest Neighbors to generate a new data point between the reference and its selected neighbor. Algorithm~\ref{alg:sintetico} presents the procedure for generating the synthetic cases using SmoteR. First the number of synthetic examples that is generated from a selected rare case, $ng$, is determined based on the percentage of over-sampling $o$ determined by the user and the dataset cardinality $|D|$ (line 3). Then, for each rare case $c$ that will be used as a reference in the generation process, its K-Nearest Neighbors are computed (Line 5) $nns$. After the set of neighbors are obtained, the algorithms execute multiple iterations to generate $ng$ synthetic examples by picking one of the examples in the $nns$ set at random and interpolating with the reference one. This generation process is presented from lines 8 to 15, which show how attribute values for the synthetic case are generated. If the attributes are numeric, the difference between the attributes of the two seed cases is calculated (line 10). Subsequently, (line 11) multiplies this difference by a random number between 0 and 1, and then adds to the example's attribute. Otherwise, a random selection between the values of the seed cases is performed. On lines 16 to 18, the value of the target is generated, calculated by the weighted average of the two cases. The weights are obtained by the distance between the new case and the two seed cases (lines 16 and 17). In~\cite{de2018utility}, this strategy is extended, and is able to handle any number of either normal or rare cases.

\begin{algorithm} [H]
    \caption{SmoteR}
    \label{alg:ST}
    
    \hspace{0.3cm} \textbf{Input:} 
     
    \hspace{0.6cm} $D$ - original dataset 
    
    \hspace{0.6cm} $t_R$ - relevance threshold 
    
    
    \hspace{0.6cm} $o$ - percentage of over-sampling 
    
    \hspace{0.6cm} $u$ - percentage of under-sampling 
    
    \hspace{0.6cm} $k$ - the number of neighbors used in case generation
    
    \hspace{0.3cm} \textbf{Output:} 
    
    \hspace{0.6cm} $newD$ - a new modiﬁed dataset
  
    \begin{algorithmic}[1]
    
    \State $Bins_R \gets \{D_i \in D: \forall(x, y) \in D_i, \phi(y) \ge t_R\}$ \Comment{partitions into relevant bins}
    \State $Bins_N \gets D \backslash Bins_R$ \Comment{partitions into normal bins}
    \State $newD \gets Bins_R$
    
    \ForEach{$B \in Bins_N$}
        \State $selNormCases \gets$ random sample of $u \times |B|$ cases of $B$ \Comment{under-sampling procedure}
        \State $newD \gets newD \bigcup selNormCases$
    \EndFor
    
    \State $newCases \gets GENSYNTHCASES(Bins_R, o, k)$  \Comment{Generation of the attribute values}
    \State $newD \gets newD \bigcup newCases$ 
    \State \Return $newD$
    
    \end{algorithmic}
    \end{algorithm}

\begin{algorithm}[H]
    \caption{Generating synthetic cases}
    \label{alg:sintetico}
    
    \hspace{0.3cm} \textbf{Input:} 
     
    \hspace{0.6cm} $D$ - original dataset
    
    \hspace{0.6cm} $o$ - percentage of over-sampling
    
    \hspace{0.6cm} $k$ - the number of neighbors used in case generation
    
    \begin{algorithmic}[1]
    \Function {GENSYNTHCASES}{$D, o, k$}

    \State $newCases \gets \{\}$
    \State  $ng \gets (o-1)\times |D|$ \Comment{number of new cases to generate for each existing case}
    \ForAll{$case \in D$}
    \State $nns \gets kNN(k, case, D_R \backslash \{case\})$ \Comment{k-Nearest Neighbours of case}
    \For {$i \gets 1$ to ng}
    \State $x \gets$ randomly choose one of the $nns$
    \ForEach{$a \in $ attributes} \Comment{generation of the attribute values}
    \If {ISNUMERIC($a$)}
        \State $diff \gets case[a] - x[a]$
        \State $new[a] \gets case[a] + RANDOM(0,1) \times diff$
    \Else
        \State $new[a] \gets$ randomly select among case$[a]$ and $x[a]$
        \EndIf
    \EndFor \Comment{generation of the target value}
    \State $d_1 \gets DIST(new, case)$
    \State $d_2 \gets DIST (new, x)$
    \State $new[y] \gets \frac{d_2 \times case[y]+d_1 \times x[y]}{d_1+d_2}$
    \EndFor
    \State $newCases \gets newCases \bigcup {new}$ \Comment{add the new synthetic case}
    \EndFor
    \State \Return $newCases$
\EndFunction
    
    \end{algorithmic}
\end{algorithm}

\subsection{Random Over-sampling}
\label{sec:RO}

The Random over-sampling~\cite{branco2019pre} strategy, presented in Algorithm~\ref{alg:RO}, works by first selecting the examples that are above the relevance threshold $t_R$ (line 2) as candidates to be duplicated, $Bins_R$. Then, for each bin $B$ belonging to the rare examples $Bins_R$, the number of replicas $tgtNr$ generated is defined according to its cardinality $|B|$ and the oversampling percentage $o$ (Line 4). The $|B|$ represents the number of elements (data points or examples) contained within that specific bin $B$. This oversampling percentage is a hyperparameter defined by the user. Random sampling is performed on line 5, and the duplicated cases are added to the new dataset ($newD$) on line 6. When performing this algorithm, no special treatment is required to generate the target values. As the examples generated are identical to the existing rare cases, the duplicated ones have exactly the same target value.

\begin{algorithm} [H]
    \caption{Random over-sampling}
    \label{alg:RO}
    
    \hspace{0.3cm} \textbf{Input:} 
    
    \hspace{0.6cm} $D$ - original dataset 
    
    \hspace{0.6cm} $t_R$ - relevance threshold 
    
    
    \hspace{0.6cm} $o$ - percentage of over-sampling
    
    \hspace{0.3cm} \textbf{Output:} 
    
    \hspace{0.6cm} $newD$ - a new modiﬁed dataset
  
    \begin{algorithmic}[1]
    \State $newD \gets D$
    \State $Bins_R \gets \{D_i \in D: \forall(x, y) \in D_i, \phi(y) \ge t_R\}$
    \ForEach{$B \in Bins_R $}
    \State $tgtNr \gets o \times |B|$ \Comment{number of replicas to be added}
    \State $selCases \gets$ sample randomly $tgtNr$ elements from $B$ 
    \State $newD \gets newD \bigcup selCases$ \Comment{add the replicas to the new data}
    \EndFor
    \State \Return $newD$
    
    \end{algorithmic}
    \end{algorithm}

\begin{algorithm}[!h]
    \caption{Random under-sampling}
    \label{alg:RU}
    
    \flushleft \hspace{0.3cm} \textbf{Input:} 
    
    \hspace{0.6cm} $D$ - original dataset 
    
    \hspace{0.6cm} $t_R$ - relevance threshold 
    
    
    \hspace{0.6cm} $u$ - percentage of under-sampling
    
    \hspace{0.3cm} \textbf{Output:} 
    
    \hspace{0.6cm} $newD$ - a new modiﬁed dataset
  
    \begin{algorithmic}[1]
    
    \State $Bins_R \gets \{D_i \in D: \forall(x, y) \in D_i, \phi(y) \ge t_R\}$
    \State $Bins_N \gets D \backslash Bins_R$
    \State $newD \gets Bins_R$
    
    \ForEach{$B \in Bins_N $}
    \State $tgtNr \gets u \times |B|$ \Comment{number of replicas to be removed}
    \State $NormCases \gets$ randomly under-sample $tgtNr$ elements from $B$ \Comment{remove the examples from B}
    \State $newD \gets newD \bigcup NormCases$
    \EndFor
    \State \Return $newD$
    
    \end{algorithmic}
    \end{algorithm}

\subsection{Random Under-sampling}
\label{sec:RU}

The Random Under-Sampling strategy (Algorithm~\ref{alg:RU}) was proposed by Torgo et al.~\cite{torgo2013smote}. In this approach, the under-sampling is performed by first using the relevance function (Section~\ref{sec:relevance}) and a relevance threshold $t_R$ to define the rare cases in the dataset (line 1). The examples below $t_R$ are considered normal, being candidates to be removed from the final dataset~\cite{branco2016ubl} (line 2), while rare cases are kept. The removal of the normal examples is thus performed according to an under-sampling rate provided by the user $u$, which defines the percentage of under-sampling applied in the dataset. For each bin $B$ belonging to the set of normal examples $Bins_N$, the number of examples removed from it is computed based on its cardinality and the percentage of undersampling $u$ (Line 5). Line 6 performs the under-sampling in $B$ by randomly selecting data points to be removed, resulting in a reduced set that is used to compose the final dataset $newD$.

\subsection{Introduction of Gaussian Noise}
\label{sec:GN}

Generating synthetic examples through Gaussian noise (Introduction of Gaussian Noise - GN) constitutes an adaptation of the method proposed in~\cite{lee1999regularization, lee2000noisy} for classification tasks to the regression context. Algorithm~\ref{alg:GN} presents the GN technique. It starts by dividing the dataset into normal cases $Bins_N$  and rare cases $Bins_R$ according to the relevance function $\phi(y)$ and the relevance threshold $t_R$ (Lines 1 and 2). Examples belonging to $Bins_N$ (i.e., normal examples) are reduced in size, using the Random under-sampling technique (lines 4 to 6). The amount of reduction is controlled by the percentage of the under-sampling hyperparameter $u$ defined by the user.

From lines 8 to 20, the over-sampling procedure is performed using the samples in $Bins_R$. For each seed case selected and used in the generation process, a total of $ng$ new artificial generated examples are added to the dataset. $ng$ is computed based on the percentage of the overs-sampling hyperparameter $o$ and the number of examples in the corresponding set $B \in Bins_R$ (Line 9). The artificial cases are generated by introducing a small perturbation on both the attributes and the target variable value of the seed case. If the attributes are nominal (line 13), the generation is performed with probability proportional to the frequency of the values found in the category (lines 14 and 15). Otherwise, for the numeric attributes, a random perturbation from a normal distribution is added, as indicated on lines 17 and 18, where $\delta$ is the perturbation amplitude defined by the user and $sd(a)$ is the standard deviation of the attribute $a$ estimated using the examples in the category. The normal perturbation is also applied to the seed target value in order to generate the target value of the newly generated example.

\begin{algorithm}[H]
    \caption{Introduction of Gaussian Noise}
    \label{alg:GN}
   
    \hspace{0.3cm} \textbf{Input:} 
    
    \hspace{0.6cm} $D$ - original dataset 
    
    \hspace{0.6cm} $t_R$ - relevance threshold 
    
    
    \hspace{0.6cm} $u$ - percentage of under-sampling 
    
    \hspace{0.6cm} $o$ - percentage of over-sampling 
    
    \hspace{0.6cm} $\delta$ - perturbation amplitude
    
    \hspace{0.3cm} \textbf{Output:} 
    
    \hspace{0.6cm} $newD$ - a new modiﬁed dataset
  
    \begin{algorithmic}[1]
    \State $Bins_N \gets 
    \{D_i \in D: \forall(x, y) \in D_i, \phi(y) < t_R\}$
    \State $Bins_R \gets 
    \{D_i \in D: \forall(x, y) \in D_i, \phi(y) \ge t_R\}$
    \State $newD \gets Bins_R$
    
    \ForEach{$B \in Bins_N $}
    \State $selNormCases \gets$ random sample of $u \times |B|$ elements from $B$
    \State $newD \gets newD \bigcup selNormCases$
    \EndFor
    
    \ForEach{$B \in Bins_R $} \Comment{over-sampling procedure}
    \State $ng \gets o \times |B|$ \Comment{number of synthetic examples for each case in \textit{B}}
    \ForEach{$case \in B$}  \Comment{generate synthetic examples}
    \For{$i \gets 1$ \textbf{to} $ng$}
    \ForEach{$a \in Attrs \bigcup Y$}
    \If{$a$ \textit{is nominal}}
    \State $probs \gets$ frequency of possible values of $a$
    \State $new[a] \gets$ sample a value from the values of $a$ with weights = $probs$
    \Else 
    \State $new[a] \gets $ multiply $case [a]$ with a random sample from \State$N(0,\delta\cdot sd(a))$
    \EndIf
    \EndFor
    \State $newD \gets newD \bigcup \{new\} $ \Comment{add synthetic case to newD}
    \EndFor
    \EndFor
    \EndFor
    \State \Return $newD$
    
    \end{algorithmic}
    \end{algorithm}

\subsection{SmoteR with Gaussian Noise}
\label{sec:SG}

The SmoteR with Gaussian Noise (SMOGN - SG)~\cite{branco2017smogn} (Algorithm~\ref{alg:SG}) combines the Random under-sampling strategy (lines 6 to 9) with two over-sampling strategies: SmoteR and Introduction of Gaussian Noise. The goal is to limit the potential risks to the SmoteR of generating bad examples when the seed and its selected neighbor are not close enough by using the more conservative strategy of just introducing Gaussian noise to generate new cases. These bad examples may not represent of the underlying data distribution and can introduce several issues like noise, bias, or inconsistencies into the dataset. Moreover, the technique aims to allow for an increase in diversity when generating examples, which is not feasible by using only the Introduction of Gaussian Noise method~\cite{branco2017smogn}. Increasing diversity means producing a comprehensive range of examples covering different data distribution aspects. The generated examples should not be overly similar or redundant. Instead, they should capture different patterns, variations, or scenarios present in the data to represent the data distribution comprehensively. Thus, SMOGN addresses the main drawbacks of SmoteR and the introduction of Gaussian noise techniques.

Line 11 determines the number of synthetic cases $ng$ that will be generated according to the percentage of the over-sampling hyperparameter $o$ and the number of existing cases in the corresponding bin $B$. Then, for each seed case in $B$, its K-Nearest Neighbors and the maximum allowed distance to generate new cases with SmoteR are computed (lines13 to 15). When the seed case and the selected neighbor are ``sufficiently near" (i.e., distance below the computed threshold $maxD$), the SMOGN generates new synthetic examples with the SmoteR (lines 17 and 18) technique. Otherwise, it uses the Introduction of Gaussian Noise method when the distance between the two examples is higher than the estimated threshold (lines 20 and 21). The generated data points are then added to the new dataset, $newD$.

\begin{algorithm}[H]
     \caption{SMOGN}
     \label{alg:SG}
     \hspace{0.3cm} \textbf{Input:} 
     
     \hspace{0.6cm} $D$ - original dataset 
     
     \hspace{0.6cm} $t_R$ - relevance threshold 
     
     \hspace{0.6cm} $u$ - percentage of under-sampling 
     
     \hspace{0.6cm} $o$ - percentage of over-sampling 
     
     \hspace{0.6cm} $k$ - number of nearest neighbors 
     
     \hspace{0.6cm} $dist$ - distance metric
    
     \hspace{0.3cm} \textbf{Output:} 
     
     \hspace{0.6cm} $newD$ - a new modiﬁed dataset
        
    \begin{algorithmic}[1]
    \State $OrdD \gets D$ order $D$ by ascending value of $Y$
    \State $\phi() \gets $ relevance function
    \State $Bins_N \gets$ partitions of consecutive examples $ \langle x_i, y_i \rangle \in OrdD$, such that $\phi(y_i) < t_R$  
     \State $Bins_R \gets$ partitions of consecutive examples $ \langle x_i, y_i \rangle \in OrdD$, such that $\phi(y_i) \ge t_R$ 
    \State $newD \gets Bins_R$ 
    
    \ForEach{$B \in Bins_N$}  \Comment{under-sampling procedure}
    \State $selNormCases \gets$ randomly
sample $u \times |B|$ cases from $B$
    \State $newD \gets newD \bigcup selNormCases$
    \EndFor
    \ForEach{$B \in Bins_R$} \Comment{over-sampling procedure}
    \State $ng \gets o \times |B|$ \Comment{number of synthetic examples for each case in \textit{B}}
    \ForEach{$case \in B$} \Comment{generate synthetic examples}
    \State $nns \gets kNN(k, case, dist)$ \Comment{K-Nearest Neighbors of case} 
    \State $DistM \gets$ distances between the case and the examples in $B$
    \State $maxD \gets median(DistM)/2$
    \For{$i \gets 1$ \textbf{to} $ng$}
    $x \gets$ randomly choose one of the $nns$
    \If {$DistM(x) < maxD$}  \Comment{safe kNN selected}
    \State $new \gets$ use SmoteR to interpolate $x$ and $case$
    \Else \Comment{non-safe kNN selected}
    \State $pert \gets min(maxD, 0.02)$
    \State $new \gets $ introduce Gaussian Noise in $case$ with a perturbation $pert$
    \EndIf
    \State $newD \gets newD \bigcup {new}$ \Comment{add synthetic case to newD}
    \EndFor
    \EndFor
    \EndFor
    \State \Return $newD$
    
    \end{algorithmic}
    \end{algorithm}

\subsection{WEighted Relevance based Combination Strategy}
\label{sec:WERCS}

The WEighted Relevance-based Combination Strategy (WERCS) strategy~\cite{branco2019pre} combines biased versions of the under- and over-sampling strategies which depend exclusively on the relevance function provided to the dataset without requiring establishing a relevance threshold. Under the WERCS, the relevance function and a modification of the relevance are used to attribute weights that are used as inclusion and removal criteria for the examples. Algorithm~\ref{alg:WERCS} details this resampling strategy. The over-sampling and under-sampling on lines 4 and 7, respectively, are performed considering weights obtained on lines 3 and 6. These weights are calculated based on the relevance function. The weights associated with over-sampling $WOver$ are proportional to the relevance function (line 3). Therefore, the higher the relevance of a case, the higher its probability of being selected for generating new cases. Conversely, the weights associated with under-sampling $WUnd$ are inversely proportional to the relevance value (line 6). Thus, normal examples, which are usually associated with lower relevance values, have a higher probability of being removed rather than used in the generation process. The number of generated and removed samples is defined based on the percentage of over-sampling $o$ and under-sampling $u$, respectively.

Therefore, the main advantage of this technique is that as a relevance threshold is not set a priori, each example can participate in both processes. Thus, both under-sampling and over-sampling strategies are applied over the entire dataset. Also, the technique eliminates the dependency on the relevance threshold $t_R$ that was a key component necessary for applying all other resampling strategies reviewed in this work.

\begin{algorithm}[]
     \caption{WEighted Relevance-based Combination Strategy (WERCS)}
     \label{alg:WERCS}
     
    \hspace{0.3cm} \textbf{Input:} 
     
    \hspace{0.6cm} $D$ - original dataset 
    
    \hspace{0.6cm} $u$ - percentage of under-sampling 
    
    \hspace{0.6cm} $o$ - percentage of over-sampling
    
    \hspace{0.3cm} \textbf{Output:} 
    
    \hspace{0.6cm} $newD$ - a new modiﬁed dataset
  
    \begin{algorithmic}[1]
    \State $\phi() \gets $ relevance function
    \State $newD \gets D$
    \State $WOver \gets \{\phi(y_i) \mid y_i \in Y\}$
    \State $Over \gets$ sample $o \times |D|$ cases from $D$ with $WOver$ weights \Comment{over-sampling procedure}
    \State $newD \gets newD \bigcup Over$
    \State $WUnd \gets \{1-\phi (y_i) \mid y_i \in Y\}$
    \State $Und \gets$ sample $u \times |D|$ cases from $D$ with $WUnd$ weights
    \State $newD \gets newD \backslash Und$ \Comment{under-sampling procedure}
    
    \State \Return $newD$
    
    \end{algorithmic}
    \end{algorithm}
    
\subsection{Advantages and Disadvantages of Strategies}
\label{sec:advantages}

\begin{figure}[]
\centering
\subfigure[Original\label{subfig:O}]{
\includegraphics[width=6cm,height=3.8cm]{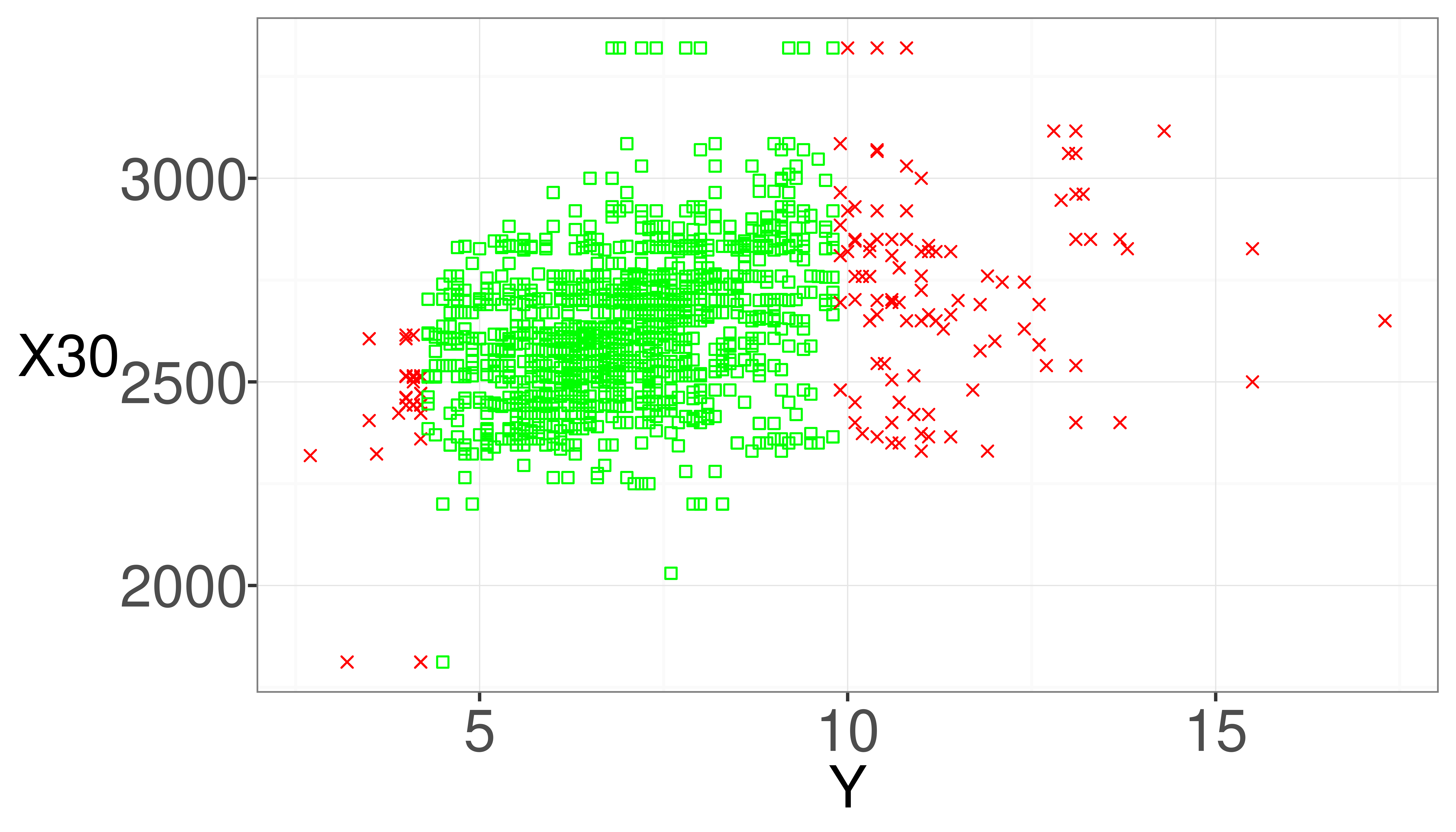}}
\subfigure[SmoteR\label{subfig:SmoteR}]{
\includegraphics[width=5.5cm,height=3.8cm]{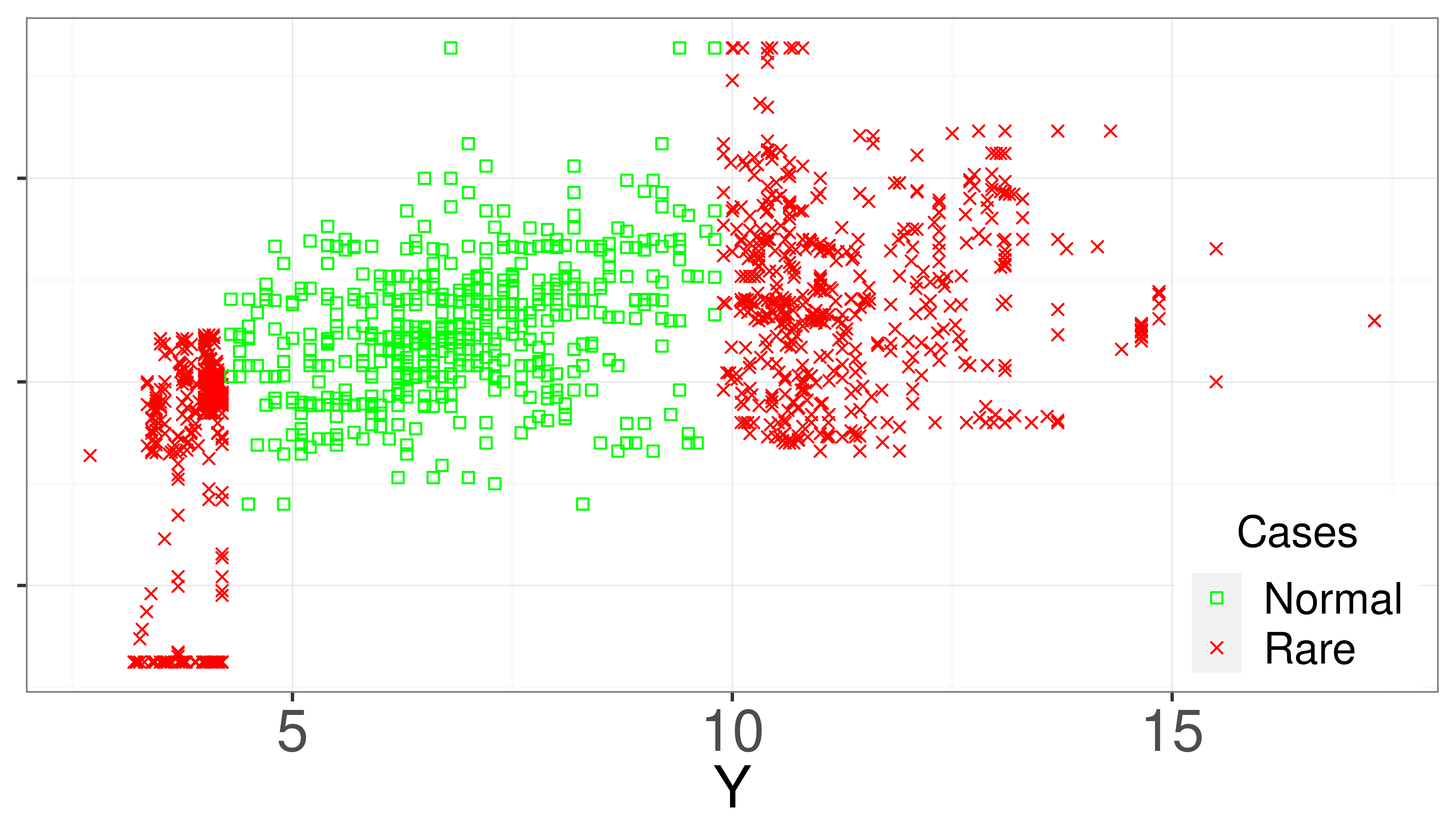}}

\subfigure[RO\label{subfig:RO}]{
\includegraphics[width=6cm,height=3.8cm]{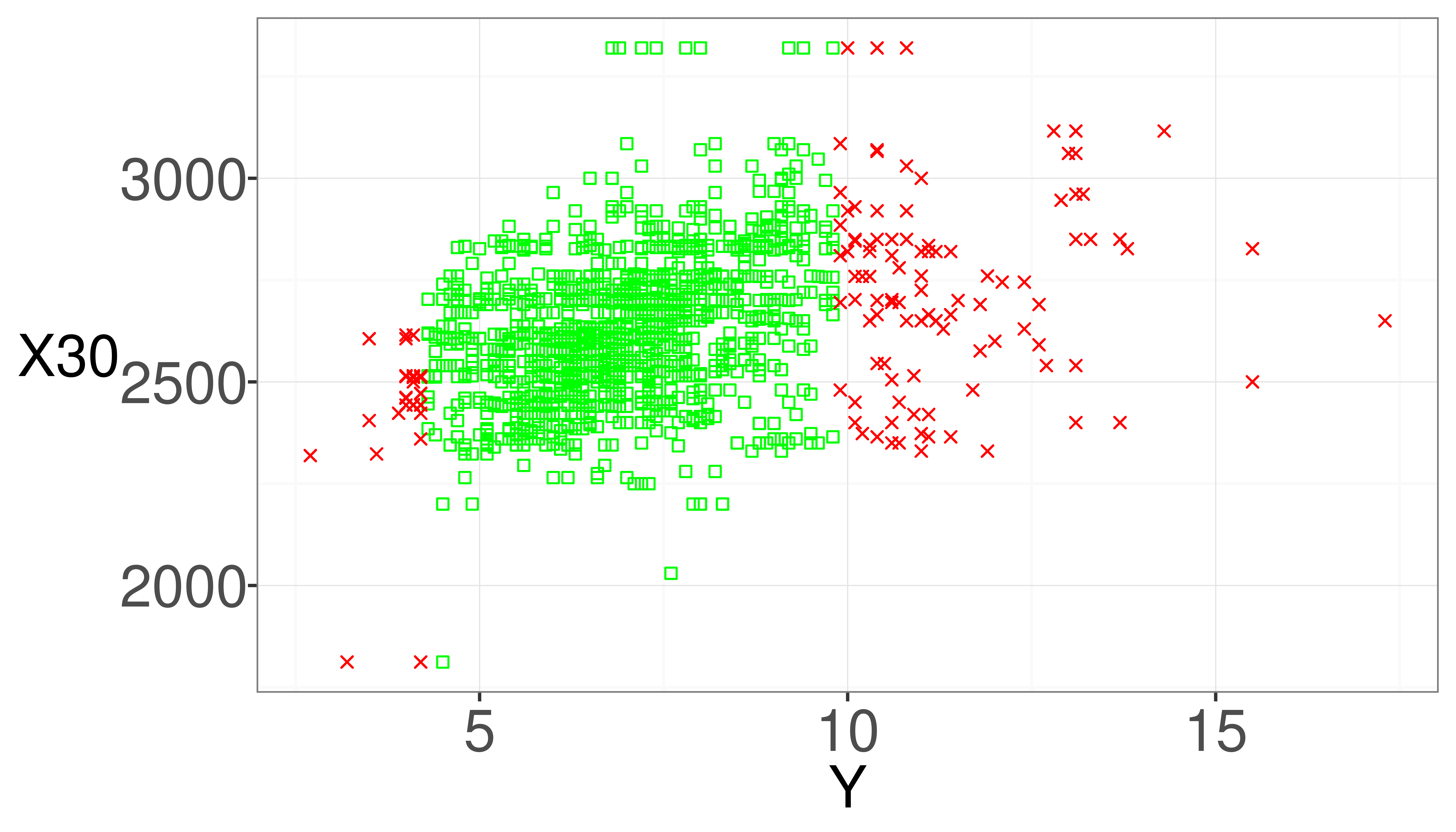}}
\subfigure[RU\label{subfig:RU}]{
\includegraphics[width=5.5cm,height=3.8cm]{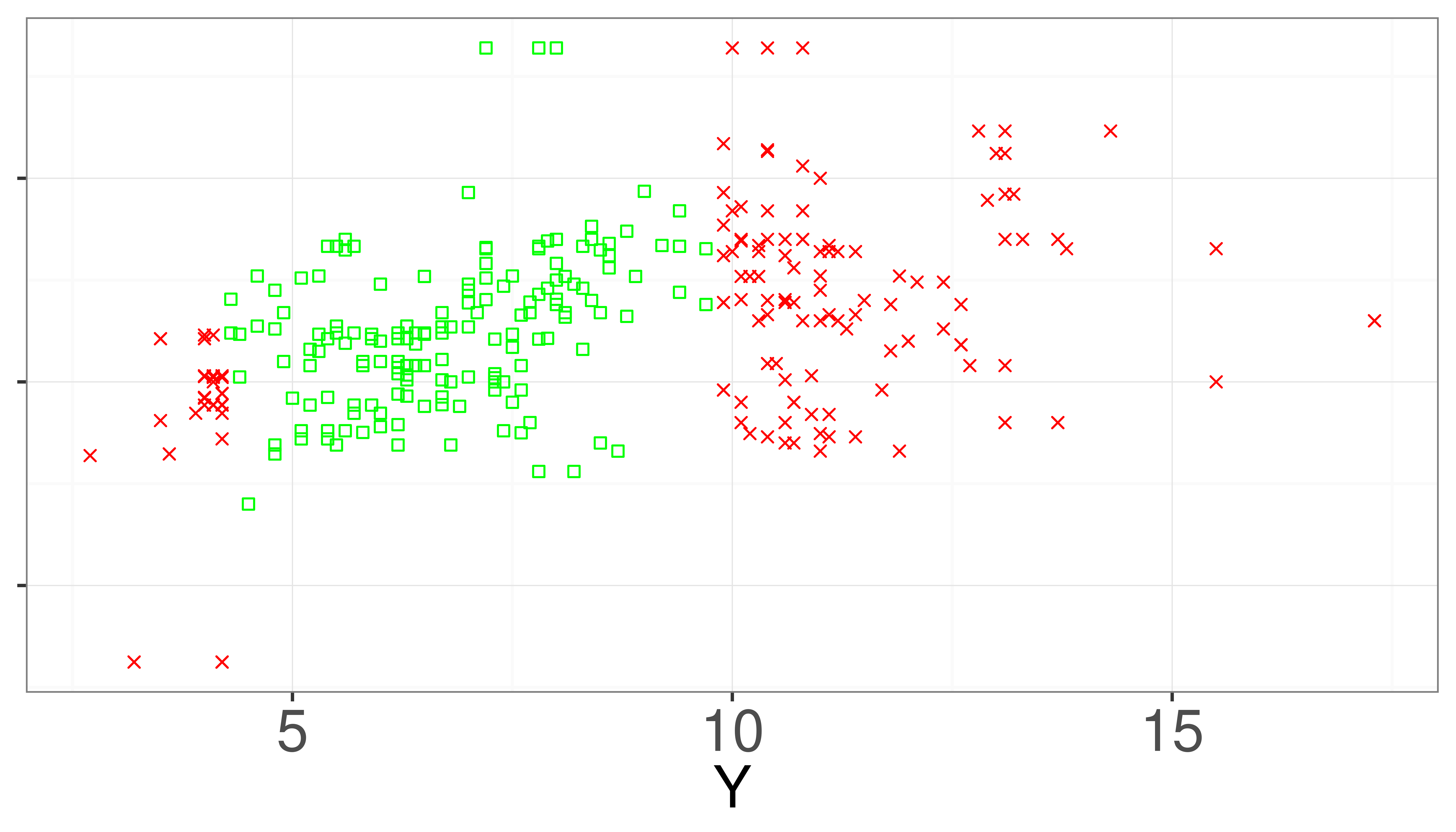}}

\subfigure[GN\label{subfig:GN}]{
\includegraphics[width=6cm,height=3.8cm]{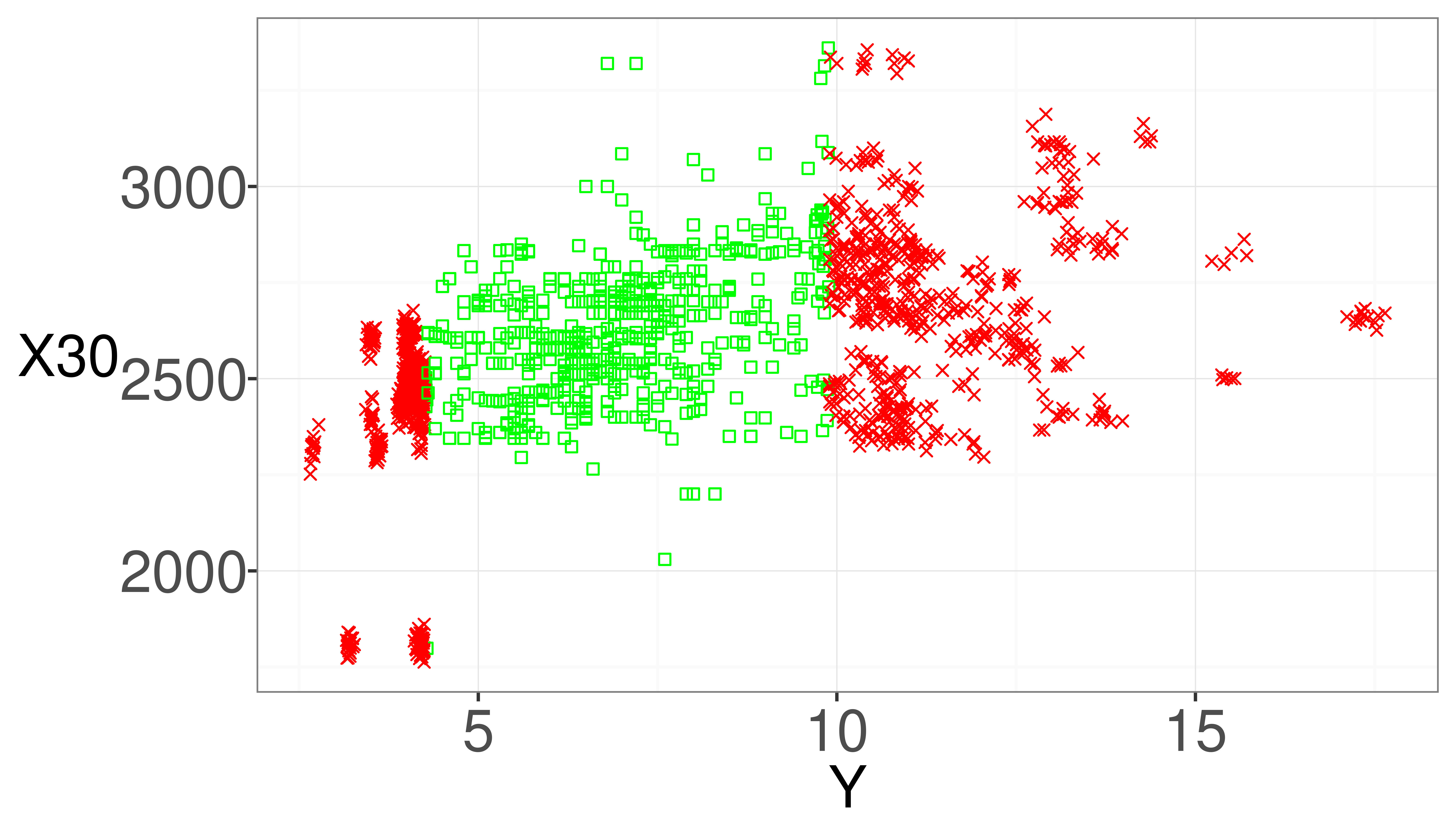}}
\subfigure[SG\label{subfig:SG}]{
\includegraphics[width=5.5cm,height=3.8cm]{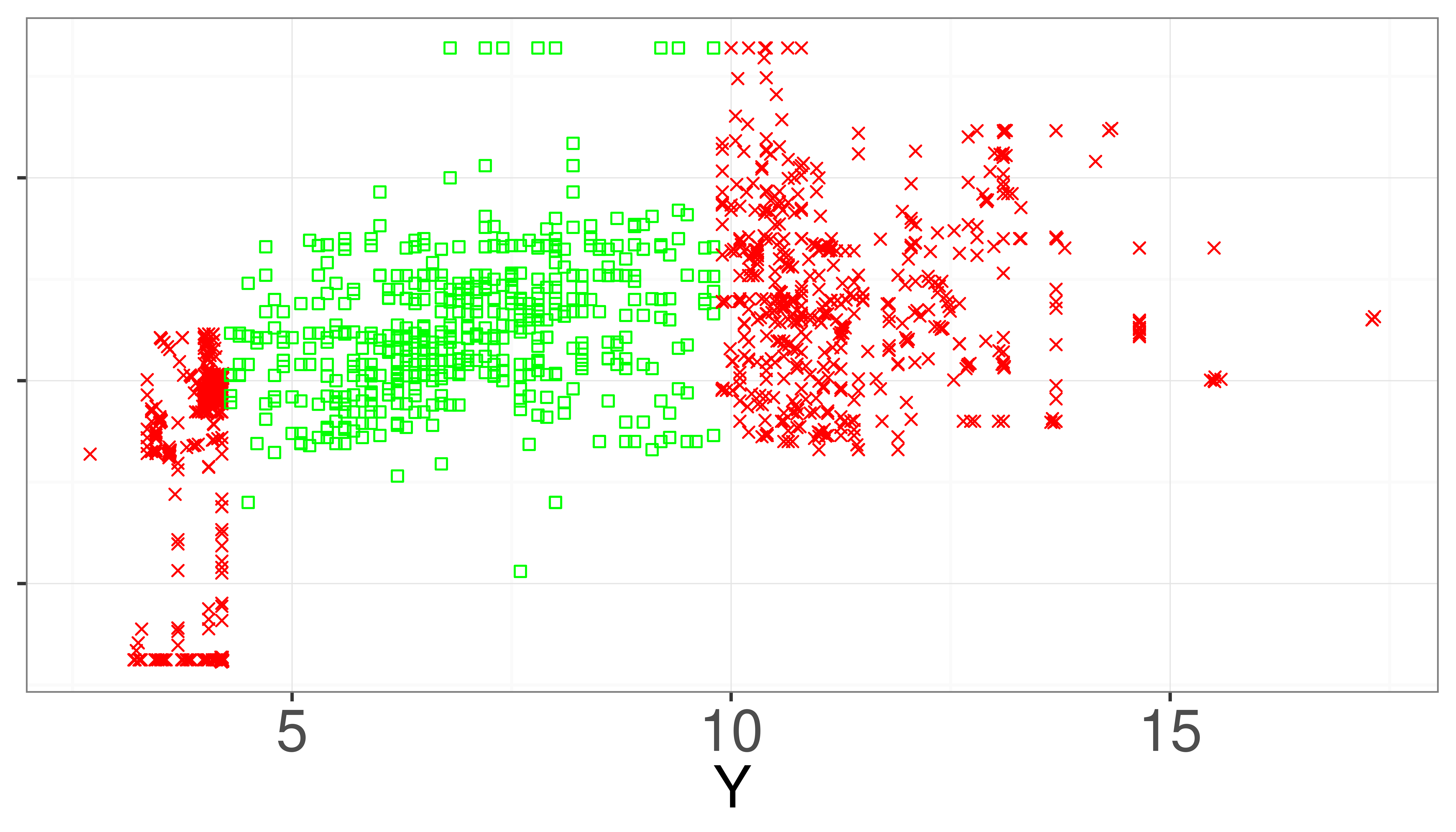}}

\subfigure[WERCS\label{subfig:WC}]{
\includegraphics[width=6cm,height=3.8cm]{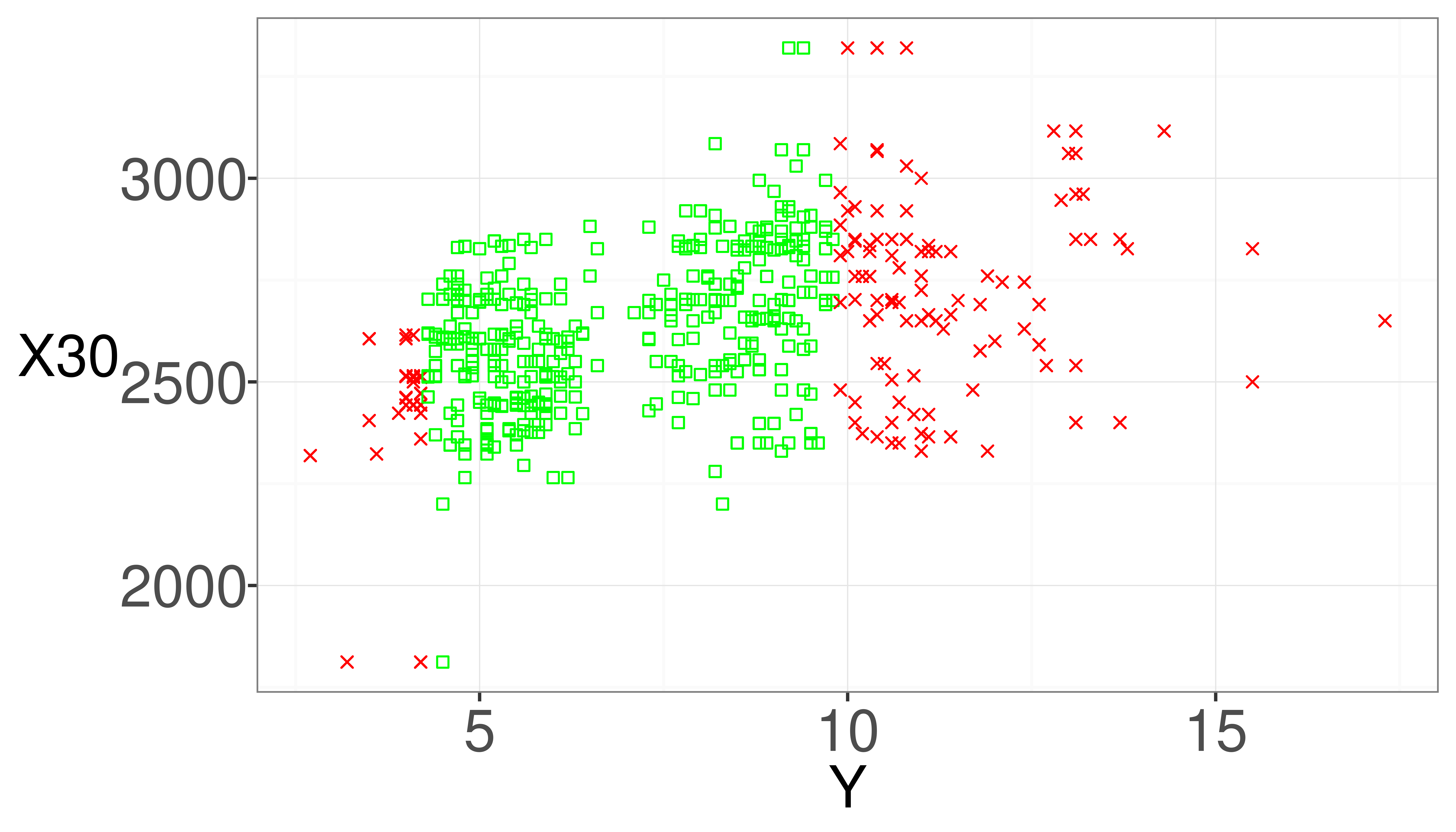}}

\caption{Distribution of the examples of the FuelCons dataset after applying the resampling strategies, considering $t_R$=0.8.}
\label{fig:strategies}
\end{figure}

The strategies to resample data can have both advantages and disadvantages. Therefore, it is crucial to understand the behavior of each strategy. While these strategies can potentially enhance learning, they can also impede the learning process of the models. Figure~\ref{fig:strategies} introduces the result of applying the resampling strategies to the \textit{FuelCons} dataset. The following values were attributed to the algorithm's parameter: $u/o$ = balance and $t_R$ = 0.8 (except for the WERCS, since it does not require establishing the threshold). The standard values were adopted for the remaining parameters. For the visualization, the target values (Y) and the attribute (X30) were considered. 

Despite selecting the nearest examples to generate new cases, SmoteR still involves the risk of the example being too far and of generating an example that does not correspond to the seed very well. This phenomenon is shown in the lower left side of Figure~\ref{subfig:SmoteR}, where the generated examples are far from the original examples. In the RO strategy, high percentages of over-sampling may cause an overfitting~\cite{branco2019pre} problem. Even though the technique increases the representation of rare cases considerably, the generated dataset does not present a high points diversity. The generation process consists in just duplicating existing samples without covering the feature space well.

Figure~\ref{subfig:RO} shows the rare data points in darker shade, given that the RO only makes copies of the examples. This can therefore lead to learning algorithms overfitting such rare examples. In addition, if the replication rate is too high, many duplicate data points are added to the dataset which can significantly increase the training time. In contrast to the RO, in the RU strategy some meaningful information may be lost due to the removal of training data (Figure~\ref{subfig:RU}), which may hamper the learning of the model. Figure~\ref{subfig:GN} shows the result after using the GN strategy, which promotes over-sampling by adding normally distributed noise. Once again, in contrast to the RO strategy, examples different from the originals ones are generated, and this diversity can help to mitigate overfitting. For the SG strategy, even though one of its goals is to reduce the risks seen in SmoteR by creating different examples from the original, Figure~\ref{subfig:SG} shows that there is still a similarity with the SmoteR distribution. However, when compared to GN, it is evident that the diversity of generated examples is higher in SG.
In the WERCS strategy (Figure~\ref{subfig:WC}), it can be seen that the green data points are divided into two groups after the under-sampling, and this result can complicate the learning process. The WERCS over-sampling strategy performs similarly to RO, where the generated data are copies of the originals; such as, no new information is added to the training set.

The advantages and disadvantages of each resampling strategy are quite evident, as is the fact that there is no perfect strategy. We hypothesize that other variables, such as the regression model and the dataset under investigation, are required to determine the best data resampling strategy. Thus, our research allows to understand the behaviors of these strategies with different regression models and problems, which in turn allows to establish directions for combinations of the three variables, namely, the resampling strategy, the regression model, and the dataset.

\section{Research methodology}
\label{sec:experiments}
\subsection{Datasets}

Experiments were performed using 30 imbalanced regression datasets chosen to match the frequency generally used in studies looking at imbalanced regression. The levels of imbalance in these datasets are defined from the relevance function (Section~\ref{sec:relevance}). A study conducted by \cite{branco2019pre} involved varying the relevance threshold from 0.5 to 1. Nevertheless, the findings showed a complex relationship between the number of rare cases, the learning algorithm, and the applied pre-processing strategy. Therefore, our experiments considered a commonly used threshold ($t_R$) of 0.8, as used in~\cite{branco2017smogn},~\cite{branco2019pre} and~\cite{branco2018rebagg}. Thus, we obtained datasets with different percentages of rare cases (imbalanced levels), varying between 5.1\% and 23.4\%. The main features of these sets are presented in Table~\ref{tab:ds}. Datasets are presented in descending order in terms of the percentage of rare cases (\%Rare). It is important to clarify that counting rare cases is conducted across the entire dataset, as commonly practiced in the literature. Counting rare cases on the entire dataset is crucial for comprehensively understanding their rarity within the data context. This approach allows us to analyze the model's behavior within the original context of the dataset. However, resampling strategies are applied only to the training set to prevent data leakage during cross-validation. The nominal attributes were codified, transforming the vector of categories into whole values between $0$ and the number of categories$-1$. As for the ordinal attributes, a pre-defined order was established (e.g., \textit{small}: 1, \textit{medium}: 2, \textit{large}: 3).

\begin{table}[ht]
\centering
\caption{Characteristics of the 30 datasets used in the experiments. N: number of cases; p.total: number of attributes; p.nom: number of nominal attributes; p.num: number of numeric attributes; nRare: number of rare  cases; Imbalance ratio (IR): $\frac{|D_R|}{|D_N|}$; \%Rare: $100 \times nRare/N$. Datasets are arranged in descending order regarding the percentage of rare cases (\%Rare).}
\begin{tabular}{llcccccc}
\hline
\textbf{Datasets} & \textbf{N}     & \textbf{p.total} & \textbf{p.nom} & \textbf{p.num} & \textbf{nRare} & \textbf{IR}    & \textbf{\%Rare} \\ \hline
wine-quality                     & 6497                      & 11      & 0     & 11    & 1523  & 0.306 & 23.4   \\
analcat-apnea3               & 450                       & 11      & 0     & 11    & 103   & 0.297 & 22.9   \\
meta                             & 528                       & 65      & 0     & 65    & 108   & 0.257 & 20.5   \\
cocomo-numeric                   & 60                        & 56      & 0     & 56    & 10    & 0.200 & 16.7   \\
Abalone                          & 4177                      & 8       & 1     & 7     & 679   & 0.194 & 16.3   \\
a3                               & 198                       & 11      & 3     & 8     & 32    & 0.193 & 16.2   \\
forestFires                      & 517                       & 12      & 0     & 12    & 79    & 0.180 & 15.3   \\
a1                               & 198                       & 11      & 3     & 8     & 28    & 0.165 & 14.1   \\
a7                               & 198                       & 11      & 3     & 8     & 27    & 0.158 & 13.6   \\
boston                           & 506                       & 13      & 0     & 13    & 65    & 0.147 & 12.8   \\
pdgfr                            & 79                        & 320     & 0     & 320   & 10    & 0.145 & 12.7   \\
sensory                          & 576                       & 11      & 0     & 11    & 69    & 0.136 & 12.0   \\
a2                               & 198                       & 11      & 3     & 8     & 22    & 0.125 & 11.1   \\
kdd-coil-1                       & 316                       & 18      & 0     & 18    & 34    & 0.121 & 10.8   \\
triazines                        & 186                       & 60      & 0     & 60    & 20    & 0.120 & 10.8   \\
airfoild                         & 1503                      & 5       & 0     & 5     & 161   & 0.120 & 10.7   \\
treasury                         & 1049                      & 15      & 0     & 15    & 109   & 0.116 & 10.4   \\
mortgage                         & 1049                      & 15      & 0     & 15    & 106   & 0.112 & 10.1   \\
debutanizer                      & 2394                      & 7       & 0     & 7     & 240   & 0.111 & 10.0   \\
fuelCons                         & 1764                      & 37      & 12    & 25    & 164   & 0.103 & 9.3    \\
heat                             & 7400                      & 11      & 3     & 8     & 664   & 0.099 & 9.0    \\
california                       & 20640                     & 8       & 0     & 8     & 1821  & 0.097 & 8.8    \\
AvailPwr                         & 1802                      & 15      & 7     & 8     & 157   & 0.095 & 8.7    \\
compactiv                        & 8192                      & 21      & 0     & 21    & 713   & 0.095 & 8.7    \\
cpuSm                            & 8192                      & 12      & 0     & 12    & 713   & 0.095 & 8.7    \\
maxTorq                          & 1802                      & 32      & 13    & 19    & 129   & 0.077 & 7.2    \\
lungcancer-shedden               & 442                       & 24      & 0     & 24    & 25    & 0.060 & 5.7    \\
space-ga                         & 3107                      & 6       & 0     & 6     & 173   & 0.059 & 5.6    \\
ConcrStr                         & 1030                      & 8       & 0     & 8     & 55    & 0.056 & 5.3    \\
Accel                            & 1732                      & 14      & 3     & 11    & 89    & 0.054 & 5.1    \\
\hline
\end{tabular}
\label{tab:ds}
\end{table}

For each dataset, the results were calculated by applying two 10-fold cross-validation repetitions (i.e., $2\times10$ cross-validation) in order to obtain the mean and standard deviation of the results. Nested cross-validation with 2-fold was employed to optimize the hyperparameters of the resampling strategies, specifically utilizing the SERA metric for optimization. The SERA metric was chosen to optimize the hyperparameters because it was specifically created for imbalanced regression. This metric evaluates models' performance in predicting extreme values, penalizing model biases without requiring a threshold, and conducting a global assessment~\cite{ribeiro2020imbalanced}. Unlike the F1-score, which conducts a local assessment by considering only rare examples, SERA evaluates all examples.


\subsection{Algorithms}

The experiments were performed with the following learning algorithms: Bagging (BG), Decision Tree (DT), Multilayer Perceptron (MLP), Random Forest (RF), Support Vector Machine (SVM), and  XGBoost (XG). Default hyperparameters were applied for these models. For details and descriptions of default hyperparameters and used packages, refer to \href{https://github.com/JusciAvelino/imbalancedRegression/blob/main/appendices/Appendix%20A.pdf}{Appendix A}.

As resample techniques, we considered the following strategies: SmoteR (SMT), Random Over-sampling (RO), Random Under-sampling (RU), Introduction of Gaussian Noise (GN), SMOGN (SG), and WEighted Relevance-based Combination Strategy (WERCS). Details about hyperparameters and packages can be found in Table~\ref{tab:algbalance}.


\begin{table}[ht]
\centering
\caption{Resampling strategies, hyperparameters, and packages used}
\begin{tabular}{lcr}
\toprule 
\textbf{Algorithms} & \textbf{Hyperparameters} & \textbf{Packages} \\
\hline  
SMT & u/o = \{balance, extreme\}, k = \{3, 5, 7\} & ImbalancedLearningRegression\tnote{1}\\ 
\hline  
RO &  o = \{balance, extreme\} & ImbalancedLearningRegression\tnote{1}\\ 
\hline  
RU &  u = \{balance, extreme\} & ImbalancedLearningRegression\tnote{1}\\

\hline  
GN                                   & \begin{tabular}[c]{@{}c@{}}u/o = \{balance, extreme\},\\ $\delta$ = \{0.00,0.05,0.10,…,0.95,1.00\}\end{tabular}                   & ImbalancedLearningRegression \tnote{1}\\
\hline  
SG                                   & \begin{tabular}[c]{@{}c@{}}u/o = \{balance, extreme\}, k = \{3, 5, 7\},\\  $\delta$ = \{0.00,0.05,0.10,…,0.95,1.00\}\end{tabular} & smogn \tnote{2}                        \\ \hline

\hline  
WERCS & u, o = \{0.3, 0.5, 0.7, 0.9\} & resreg\tnote{3}\\ 

\bottomrule 
\end{tabular}
\begin{tablenotes}

\item[1] \url{https://pypi.org/project/ImbalancedLearningRegression/} - Version 0.0.1
\item[2] \url{https://pypi.org/project/smogn/} - Version 0.1.2
\item[3] \url{https://pypi.org/project/resreg/} - Version 0.2

\end{tablenotes}
\label{tab:algbalance}
\end{table}

\subsection{Model Evaluation}

In imbalanced tasks, choosing the appropriate metrics for model evaluation is essential. This work uses the F1-score and SERA metrics to evaluate regression models, allowing the evaluation of different perspectives of the model performance. While the F1-score metric is based on the concept of utility-based evaluation and performs a local assessment according to the definition of a relevance threshold, the SERA metric evaluates the effectiveness of models in predicting extreme values while penalizing several model biases without the need for a threshold, and performing a global assessment~\cite{ribeiro2020imbalanced}. The results for the RMSE and MAE metrics can be consulted in the supplementary material  \href{https://github.com/JusciAvelino/imbalancedRegression/blob/main/appendices/Appendix%20B.pdf}{(Appendix B)} for benchmarking purposes.

\section{Results}
\label{sec:results}

The experiments aimed at answering the following research questions:  

\begin{enumerate}
  \item Is it worth using resampling strategies?
  \item Which resampling strategies influence the predictive performance the most?
  \item Does the choice of best strategy depend on the problem, the learning model, and the metrics used?
  \item Does the number of training examples resulting from each strategy influence the results?
  \item Do the features of the data (percentage of rare cases, number of rare cases, dataset size, number of attribues and imbalance ratio) impact the predictive performance of the models?
\end{enumerate}

Tables~\ref{tab:venceuf1}~and~\ref{tab:venceusera} show how many times each algorithm obtains the highest value for the F1-score and SERA metrics, respectively. Where there is a tie, each of the $n$ tied strategies receives $1/n$ point. Each row in this table must add up to 30, the number of datasets assessed. For both metrics used, we found that the larger number of wins occurs when using some of the resampling strategies, which points to an advantage of using such strategies, and highlighting the RO and GN considering the F1-score, and the GN and WERCS, according to SERA. Another point observed is that the choice of best strategy possibly depends on the regression model used. As for the metrics, both agree regarding the GN strategy. By observing the score by rows, also in Tables~\ref{tab:venceuf1} and~\ref{tab:venceusera}, it is clear that there is no general agreement between the datasets for a resampling strategy since each point is a dataset, and all of them are distributed in different strategies. The results per learning algorithm, including mean and standard deviation, can be accessed in the supplementary material --  Tables~\ref{tab:venceuf1}~and~\ref{tab:venceusera} show how many times each algorithm obtains the highest value for the F1-score and SERA metrics, respectively. Where there is a tie, each of the $n$ tied strategies receives $1/n$ point. Each row in this table must add up to 30, the number of datasets assessed. For both metrics used, we found that the larger number of wins occurs when using some of the resampling strategies, which points to an advantage of using such strategies, and highlighting the RO and GN considering the F1-score, and the GN and WERCS, according to SERA. Another point observed is that the choice of best strategy possibly depends on the regression model used. As for the metrics, both agree regarding the GN strategy. By observing the score by rows, also in Tables~\ref{tab:venceuf1} and~\ref{tab:venceusera}, it is clear that there is no general agreement between the datasets for a resampling strategy since each point is a dataset, and all of them are distributed in different strategies. The results per learning algorithm, including mean and standard deviation, can be accessed in the supplementary material --  \href{https://github.com/JusciAvelino/imbalancedRegression/blob/main/appendices/Appendix%20B.pdf}{Appendix B}..





\begin{table}[h]
\centering
\caption{Number of times each algorithm and resampling strategy achieved the best result according to the F1-score metric.}
\label{tab:venceuf1}
\begin{tabular}{@{}cccccccc@{}}
\hline
      & None & SMT & RO          & RU & GN            & SG   & WERCS \\ \hline
BG                       & 3                                                & 3                       & 13                     & 2                      & 6                      & 3                      & 0                         \\
DT                       & 6                                                & 5                       & 8                      & 3                      & 6                      & 2                      & 0                         \\
MLP                      & 3                                                & 4                       & 12                     & 1                      & 3                      & 6                      & 1                         \\
RF                       & 2                                                & 4                       & 9                      & 6                      & 3                      & 3                      & 3                         \\
SVR                      & 1                                                & 5                       & 13                     & 1                      & 2                      & 6                      & 2                         \\
XG                       & 4.7                                              & 1                       & 7                      & 5                      & 5.2                    & 3.2                    & 4                         \\ \hline
Total                    & 19.7                                             & 22                      & \textbf{62}            & 18                     & \textbf{25.2}          & 23.2                   & 10            \\ \hline
\end{tabular}
\end{table}



\begin{table}[h]
\centering
\caption{Number of times each algorithm and resampling strategy reached the best result according to the SERA metric.}
\label{tab:venceusera}
\begin{tabular}{@{}cccccccc@{}}
\hline
      & None & SMT & RO & RU & GN            & SG   & WERCS       \\ \hline
BG                       & 4                                                & 3                       & 3                      & 5                      & 4                      & 1                      & 10                        \\
DT                       & 4                                                & 2                       & 7                      & 6                      & 5                      & 2                      & 4                         \\
MLP                      & 2                                                & 4                       & 9                      & 1                      & 4                      & 7                      & 3                         \\
RF                       & 2                                                & 1                       & 4                      & 7                      & 6                      & 2                      & 8                         \\
SVR                      & 0                                                & 5                       & 1                      & 3                      & 12                     & 5                      & 4                         \\
XG                       & 3                                                & 1                       & 2                      & 5                      & 8.5                    & 3.5                    & 7                         \\ \hline
Total                    & 15                                               & 16                      & 26           & 27                     & \textbf{39.5}          & 20.5                   & \textbf{36}               \\ \hline
\end{tabular}
\end{table}


To identify the best way to preprocess each dataset, Tables~\ref{tab:mpf1} and~\ref{tab:mpsera} introduce the best and worst results for the F1-score and SERA metrics, respectively. The results show that most datasets have distinct preferences in terms of combining the best learning model and the resampling strategy. This distinction is also found for the metrics used. As for the worst results, the SVR and MLP, without preprocessing, is the worst combinations for both metrics. Thus, balancing the dataset before applying these models is crucial to reaching more promising results. It is also crucial to note the significant difference between the best and worst results per problem. So, obtaining good results depends on the correct choice of resampling strategy and learning model. Unfortunately, the SG strategy failed to perform on the california, heat, and wine-quality datasets. These are large datasets, highlighting the potential challenges in optimizing hyperparameters, rendering the use of this model impractical.


\begin{table}[ht]
\centering
\caption{Best and worst results for each dataset based on the F1-score metric}
\label{tab:mpf1}
\begin{tabular}{@{}lrlrl@{}}
\toprule
\textbf{Datasets} & \multicolumn{2}{c}{\textbf{Best result}} & \multicolumn{2}{c}{\textbf{Worst result}} \\ \midrule
wine-quality                                               & 0.738                     & RF.RO                       & 0.000              & SG\tnote{*}         \\
analcat-apnea3                                         & 0.233                     & MLP.NONE                    & 2.00e-5        & SVR.NONE        \\
meta                                                       & 0.435                     & BG.GN                       & 2.00e-5        & SVR.NONE        \\
cocomo-numeric                                             & 0.371                     & RF.RU                       & 1.60e-5        & SVR.NONE        \\
Abalone                                                    & 0.702                     & RF.RU                       & 1.79e-1        & SVR.NONE        \\
a3                                                         & 0.506                     & RF.RU                       & 1.95e-5        & SVR.NONE        \\
forestFires                                                & 0.403                     & SVR.SMT                     & 2.00e-5        & SVR.NONE        \\
a1                                                         & 0.723                     & RF.SG                       & 2.00e-5        & SVR.NONE        \\
a7                                                         & 0.411                     & SVR.RO                      & 2.00e-5        & SVR.NONE        \\
boston                                                     & 0.893                     & RF.RO                       & 2.00e-5        & SVR.NONE        \\
pdgfr                                                      & 0.229                     & RF.GN                       & 1.60e-5        & RF.SG           \\
sensory                                                    & 0.672                     & XG.GN                       & 2.00e-5        & SVR.NONE        \\
a2                                                         & 0.580                     & RF.RU                       & 2.00e-5        & SVR.NONE        \\
kdd-coil-1                                                 & 0.677                     & RF.RU                       & 1.90e-5        & SVR.NONE        \\
triazines                                                  & 0.226                     & RF.WERCS                    & 0.028          & RF.NONE         \\
airfoild                                                   & 0.951                     & BG.RO                       & 2.00e-5        & SVR.WERCS       \\
treasury                                                   & 0.980                     & RF.GN                       & 0.777          & SVR.NONE        \\
mortgage                                                   & 0.985                     & RF.RO                       & 0.834          & SVR.NONE        \\
debutanizer                                                & 0.901                     & RF.SMT                      & 0.646          & MLP.GN          \\
fuelCons                                                   & 0.942                     & XG.GN                       & 0.094          & MLP.RU          \\
heat                                                       & 0.989                     & XG.RO                       & 0.000              & SG\tnote{*}        \\
california                                                 & 0.902                     & XG.NONE                     & 0.000              & SG\tnote{*}         \\
AvailPwr                                                   & 0.977                     & XG.NONE                     & 0.725          & SVR.NONE        \\
compactiv                                                  & 0.528                     & RF.SG                       & 0.109          & MLP.RO          \\
cpuSm                                                      & 0.526                     & RF.SMT                      & 0.105          & MLP.RO          \\
maxTorq                                                    & 0.988                     & XG.WERCS                    & 2.00e-5        & SVR.NONE        \\
lungcancer-shedden                                         & 0.665                     & MLP.SG                      & 2.00e-5        & SVR.NONE        \\
space-ga                                                   & 0.802                     & XG.GN                       & 2.00e-5        & SVR.NONE        \\
ConcrStr                                                   & 0.966                     & XG.GN                       & 2.00e-5        & SVR.RU          \\
Accel                                                      & 0.961                     & XG.RO                       & 2.00e-5        & SVR.GN     
\\ \bottomrule
\end{tabular}

\begin{tablenotes}

\item[*] not completed in a reasonable timeframe

\end{tablenotes}

\end{table}


\begin{table}[]
\centering
\caption{Best and worst results for each dataset based on the SERA metric}
\label{tab:mpsera}
\begin{tabular}{@{}lrlrl@{}}
\toprule
\textbf{Datasets} & \multicolumn{2}{c}{\textbf{Best result}} & \multicolumn{2}{c}{\textbf{Worst result}} \\ \midrule
wine-quality                                               & 1.43e+2                    & RF.WERCS                   & 0.000               & SG\tnote{*}        \\
analcat-apnea3                                         & 3.93e+7                    & RF.GN                      & 4.80e+8         & SVR.NONE       \\
meta                                                       & 2.79e+7                    & MLP.GN                     & 3.83e+7         & DT.SG          \\
cocomo-numeric                                             & 4.16e+5                    & XG.SMT                     & 2.91e+6         & SVR.NONE       \\
Abalone                                                    & 1.17e+3                    & RF.WERCS                   & 2.59e+3         & SVR.NONE       \\
a3                                                         & 3.30e+2                    & SVR.GN                     & 2.48e+7         & MLP.RU         \\
forestFires                                                & 1.88e+5                    & MLP.SG                     & 3.02e+5         & DT.GN          \\
a1                                                         & 1.63e+3                    & SVR.SG                     & 1.76e+6         & MLP.NONE       \\
a7                                                         & 2.58e+2                    & XG.RU                      & 2.62e+7         & MLP.RU         \\
boston                                                     & 2.30e+2                    & BG.WERCS                   & 2.51e+8         & MLP.RU         \\
pdgfr                                                      & 3.86e-2                    & RF.GN                      & 2.40e-1         & DT.WERCS       \\
sensory                                                    & 1.39e+1                    & RF.RU                      & 1.16e+2         & MLP.SMT        \\
a2                                                         & 6.91e+2                    & SVR.GN                     & 2.28e+7         & MLP.RU         \\
kdd-coil-1                                                 & 2.43e+3                    & SVR.RU                     & 9.09e+3         & SVR.NONE       \\
triazines                                                  & 6.99e-2                    & RF.RU                      & 3.54e-1         & MLP.NONE       \\
airfoild                                                   & 6.78e+7                    & XG.RO                      & 2.05e+11        & SVR.NONE       \\
treasury                                                   & 3.08e+0                    & RF.GN                      & 1.78e+3         & MLP.RU         \\
mortgage                                                   & 1.30e+0                    & XG.RO                      & 1.70e+2         & DT.SG          \\
debutanizer                                                & 3.23e-1                    & RF.SG                      & 2.52e+0         & MLP.NONE       \\
fuelCons                                                   & 1.52e+1                    & XG.GN                      & 1.32e+7         & MLP.RU         \\
heat                                                       & 6.42e+2                    & XG.GN                      & 0.000               & SG\tnote{*}       \\
california                                                 & 1.87e+12                   & RF.RU                      & 0.000               & SG\tnote{*}       \\
AvailPwr                                                   & 4.72e+3                    & XG.NONE                    & 1.77e+5         & MLP.RU         \\
compactiv                                                  & 1.61e+3                    & RF.WERCS                   & 6.26e+7         & MLP.NONE       \\
cpuSm                                                      & 2.11e+3                    & RF.WERCS                   & 5.25e+7         & MLP.RO         \\
maxTorq                                                    & 4.63e+3                    & XG.GN                      & 1.17e+6         & SVR.NONE       \\
lungcancer-shedden                                         & 5.75e+1                    & SVR.GN                     & 2.16e+2         & SVR.NONE       \\
space-ga                                                   & 1.78e+0                    & XG.WERCS                   & 4.16e+10        & MLP.RO         \\
ConcrStr                                                   & 5.64e+2                    & XG.WERCS                   & 9.27e+3         & SVR.GN         \\
Accel                                                      & 2.34e+1                    & XG.GN                      & 7.41e+5         & MLP.RU
      \\ \bottomrule
\end{tabular}
\begin{tablenotes}

\item[*] not completed in a reasonable timeframe

\end{tablenotes}
\end{table}


We applied the Friedman test to better measure the advantage of using resampling strategies ($p-value <0.05$). The Friedman statistical test was chosen since it can compare multiple techniques over several datasets~\cite{demvsar2006statistical}. For this measurement, ranking sequences are compared. Tables~\ref{tab:rankF1} and~\ref{tab:rankSERA} present the mean ranking of the means of the algorithms with a combination of each resampling strategy, considering the F1-score and SERA metrics. The lower the ranking, the better the algorithm performance. The algorithms used present significant differences. In general, the best average rank of each algorithm was obtained by using some of the resampling strategies evaluated in this work.

\begin{center}

\begin{table}[h]
    \begin{minipage}{0.45\linewidth}
    \centering     
        \caption{Average ranking (F1-score).}
        \label{tab:rankF1}
        \scalefont{0.9}
        \begin{tabular}{@{}lr@{}}
        \toprule
        \textbf{Algorithm} & \textbf{Average} \\ \midrule
RF.RO                               & 10.0                              \\
RF.GN                               & 11.8                              \\
RF.RU                               & 11.9                              \\
XG.RU                               & 12.0                              \\
BG.RO                               & 12.2                              \\
RF.WERCS                            & 12.4                              \\
XG.WERCS                            & 12.8                              \\
XG.GN                               & 13.3                              \\
BG.RU                               & 13.6                              \\
XG.RO                               & 14.6                              \\
BG.WERCS                            & 15.5                              \\
XG.NONE                             & 15.6                              \\
BG.GN                               & 16.0                              \\
RF.NONE                             & 17.3                              \\
RF.SMT                              & 17.8                              \\
XG.SMT                              & 18.4                              \\
XG.SG                               & 18.5                              \\
BG.NONE                             & 18.8                              \\
BG.SMT                              & 18.8                              \\
DT.RO                               & 19.3                              \\
RF.SG                               & 20.3                              \\
DT.NONE                             & 20.4                              \\
BG.SG                               & 20.5                              \\
DT.RU                               & 20.9                              \\
DT.GN                               & 21.0                              \\
DT.WERCS                            & 21.1                              \\
DT.SMT                              & 22.9                              \\
DT.SG                               & 25.1                              \\
SVR.RO                              & 26.5                              \\
SVR.SMT                             & 26.7                              \\
MLP.RO                              & 26.8                              \\
SVR.GN                              & 27.6                              \\
SVR.SG                              & 28.4                              \\
MLP.SMT                             & 29.8                              \\
MLP.GN                              & 30.8                              \\
SVR.WERCS                           & 31.3                              \\
MLP.WERCS                           & 31.7                              \\
MLP.SG                              & 31.7                              \\
MLP.NONE                            & 32.5                              \\
MLP.RU                              & 33.3                              \\
SVR.RU                              & 33.8                              \\
SVR.NONE                            & 39.6                                 \\  \bottomrule
        \end{tabular}
    \end{minipage}
    \hfill
    \begin{minipage}{0.25\linewidth}
        \caption{Average ranking (SERA).}
        \label{tab:rankSERA}
        \scalefont{0.9}
        \begin{tabular}{@{}lr@{}}
        \toprule
        \textbf{Algorithm} & \textbf{Average} \\ \midrule
RF.RU                               & 8.7                               \\
RF.GN                               & 10.0                              \\
RF.WERCS                            & 10.9                              \\
RF.RO                               & 11.3                              \\
XG.GN                               & 11.5                              \\
XG.WERCS                            & 12.4                              \\
XG.RU                               & 12.5                              \\
BG.WERCS                            & 13.4                              \\
RF.NONE                             & 13.6                              \\
XG.SG                               & 13.6                              \\
XG.NONE                             & 13.9                              \\
BG.RU                               & 14.2                              \\
RF.SG                               & 14.3                              \\
BG.GN                               & 14.7                              \\
XG.RO                               & 14.8                              \\
BG.RO                               & 15.4                              \\
RF.SMT                              & 15.8                              \\
XG.SMT                              & 17.8                              \\
BG.NONE                             & 18.3                              \\
BG.SG                               & 18.7                              \\
BG.SMT                              & 19.6                              \\
SVR.SG                              & 23.9                              \\
SVR.GN                              & 23.9                              \\
DT.GN                               & 26.8                              \\
DT.NONE                             & 26.9                              \\
SVR.RO                              & 27.1                              \\
DT.RU                               & 27.2                              \\
DT.RO                               & 27.3                              \\
DT.SG                               & 27.4                              \\
SVR.RU                              & 27.5                              \\
MLP.SG                              & 27.8                              \\
SVR.SMT                             & 28.0                              \\
DT.WERCS                            & 28.6                              \\
MLP.RO                              & 28.7                              \\
SVR.WERCS                           & 28.9                              \\
MLP.GN                              & 29.4                              \\
MLP.SMT                             & 30.4                              \\
MLP.WERCS                           & 30.7                              \\
DT.SMT                              & 31.4                              \\
MLP.NONE                            & 33.9                              \\
MLP.RU                              & 35.1                              \\
SVR.NONE                            & 36.6                                   \\ \bottomrule
\end{tabular}
\end{minipage}
\end{table}

\end{center}

\begin{figure}[h]
\subfigure[Results of the F1-score metric for the BG algorithm\label{CDDT}]{
\includegraphics[width=6.5cm,height=3.5cm]{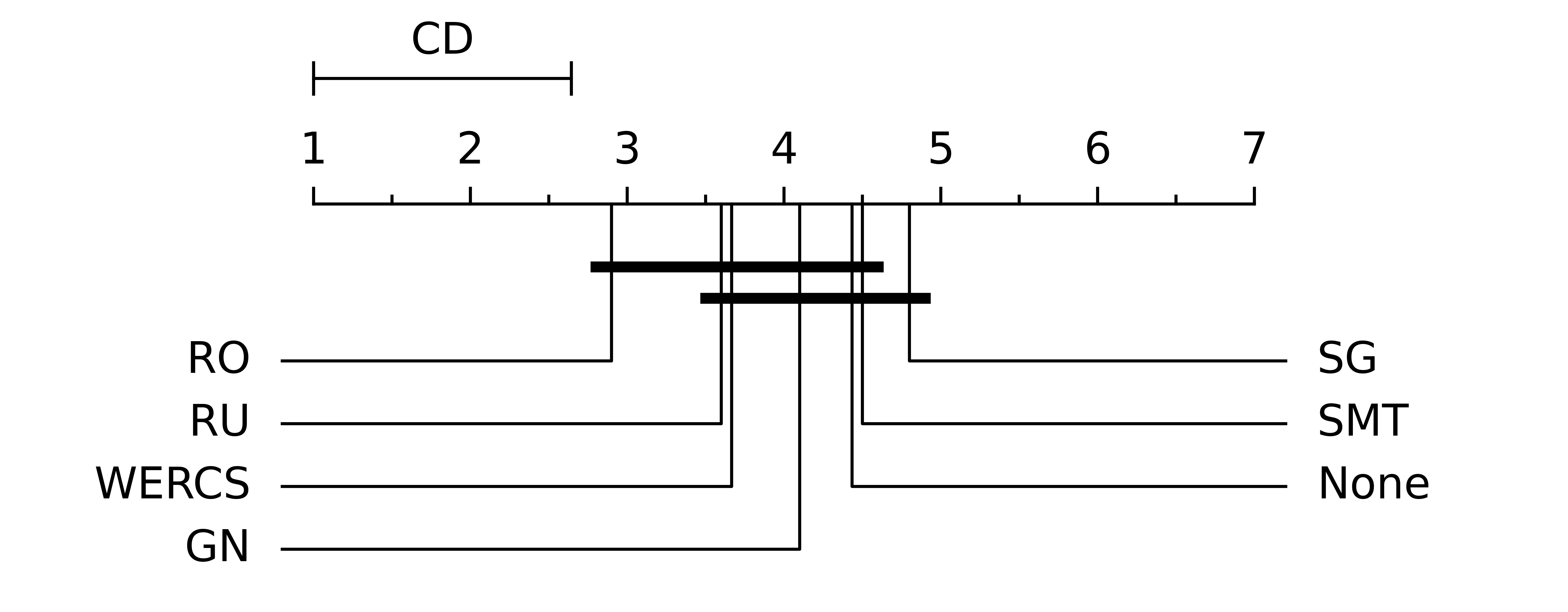}}
\subfigure[Results of the F1-score metric for the DT algorithm\label{CDBG}]{
\includegraphics[width=6.5cm,height=3.5cm]{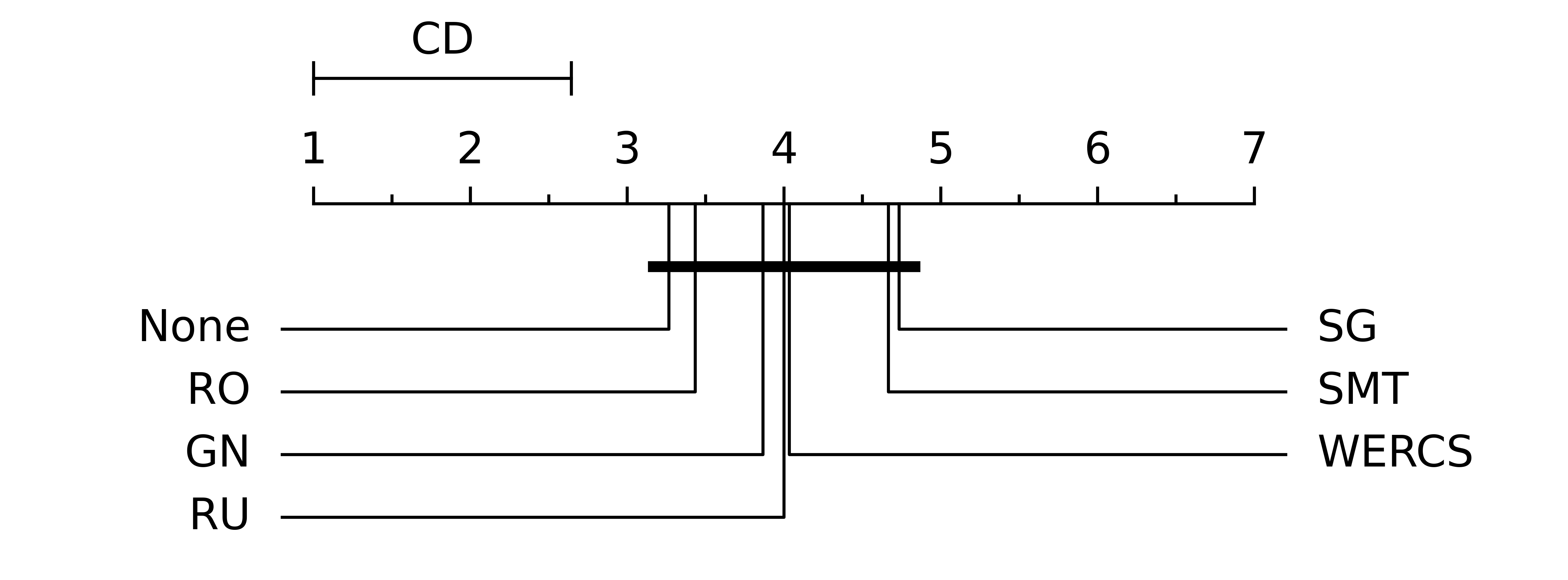}}
\subfigure[Results of the F1-score metric for the MLP algorithm\label{CDXG}]{
\includegraphics[width=6.5cm,height=3.5cm]{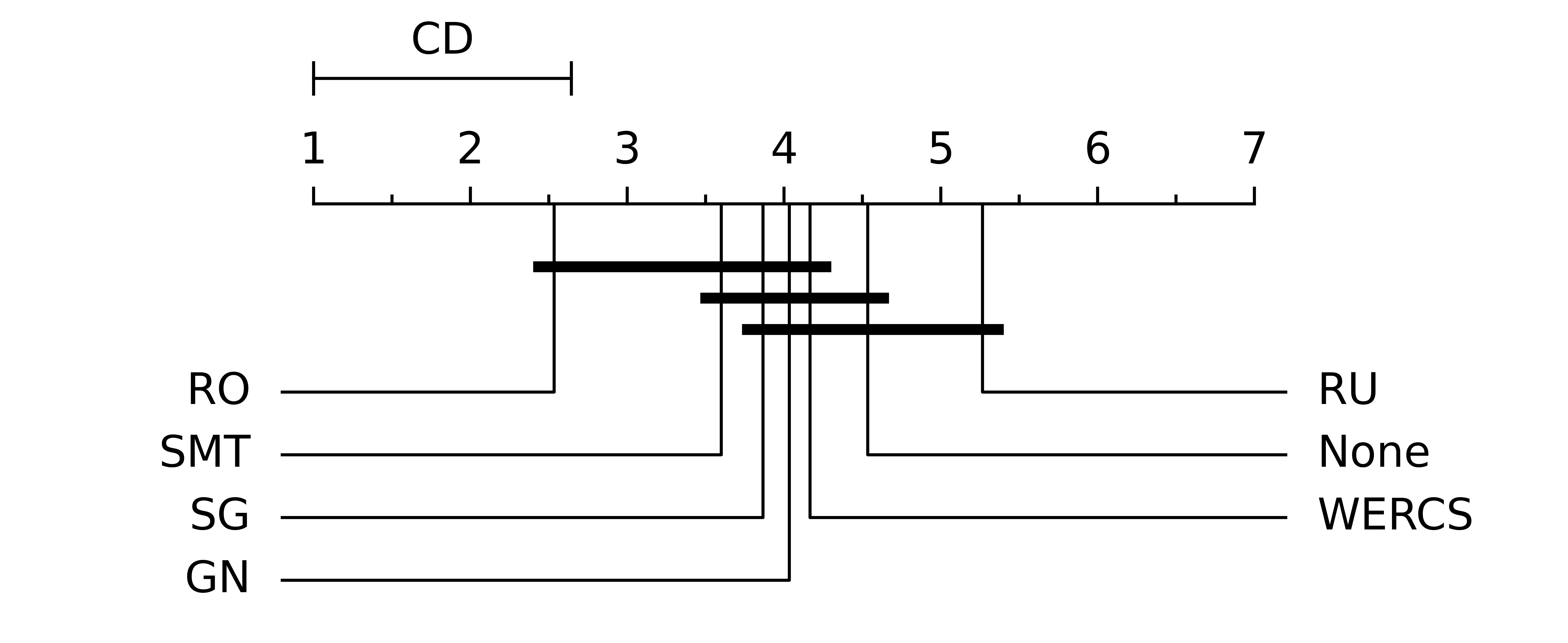}}
\subfigure[Results of the F1-score metric for the RF algorithm\label{CDRF}]{
\includegraphics[width=6.5cm,height=3.5cm]{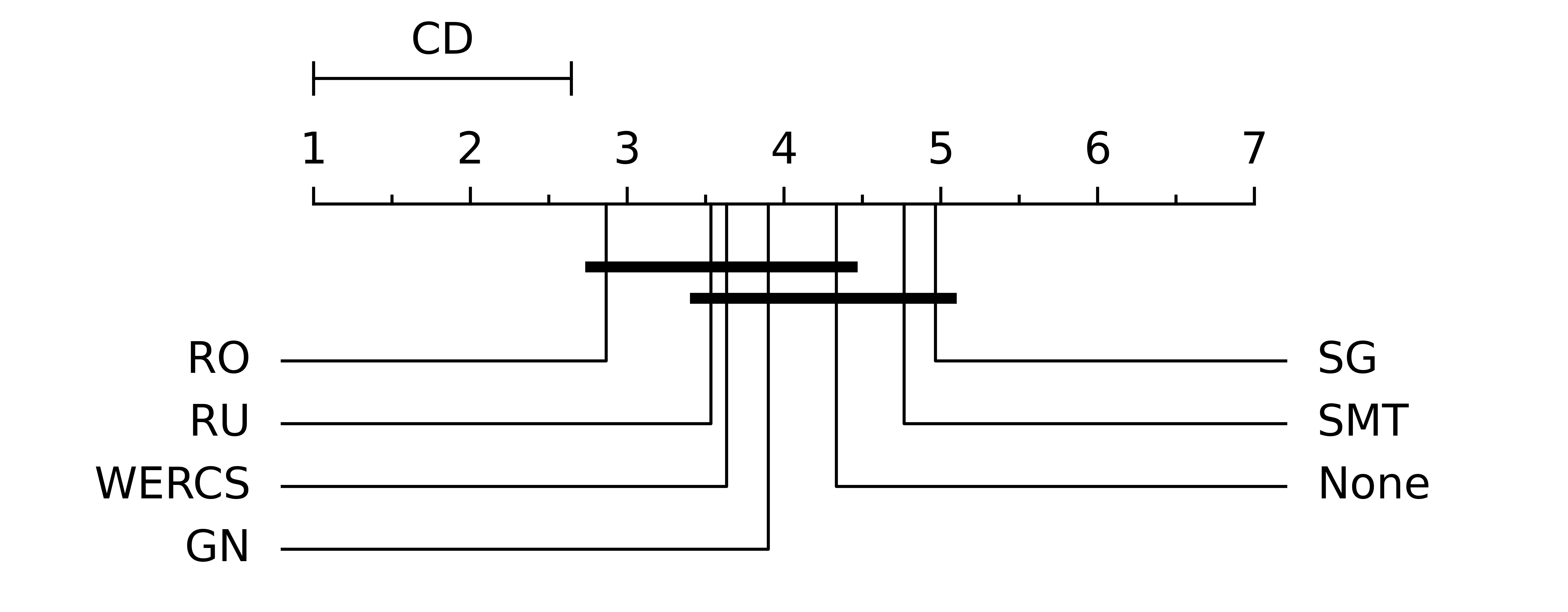}}
\subfigure[Results of the F1-score metric for the SVR algorithm\label{CDSVM}]{
\includegraphics[width=6.5cm,height=3.5cm]{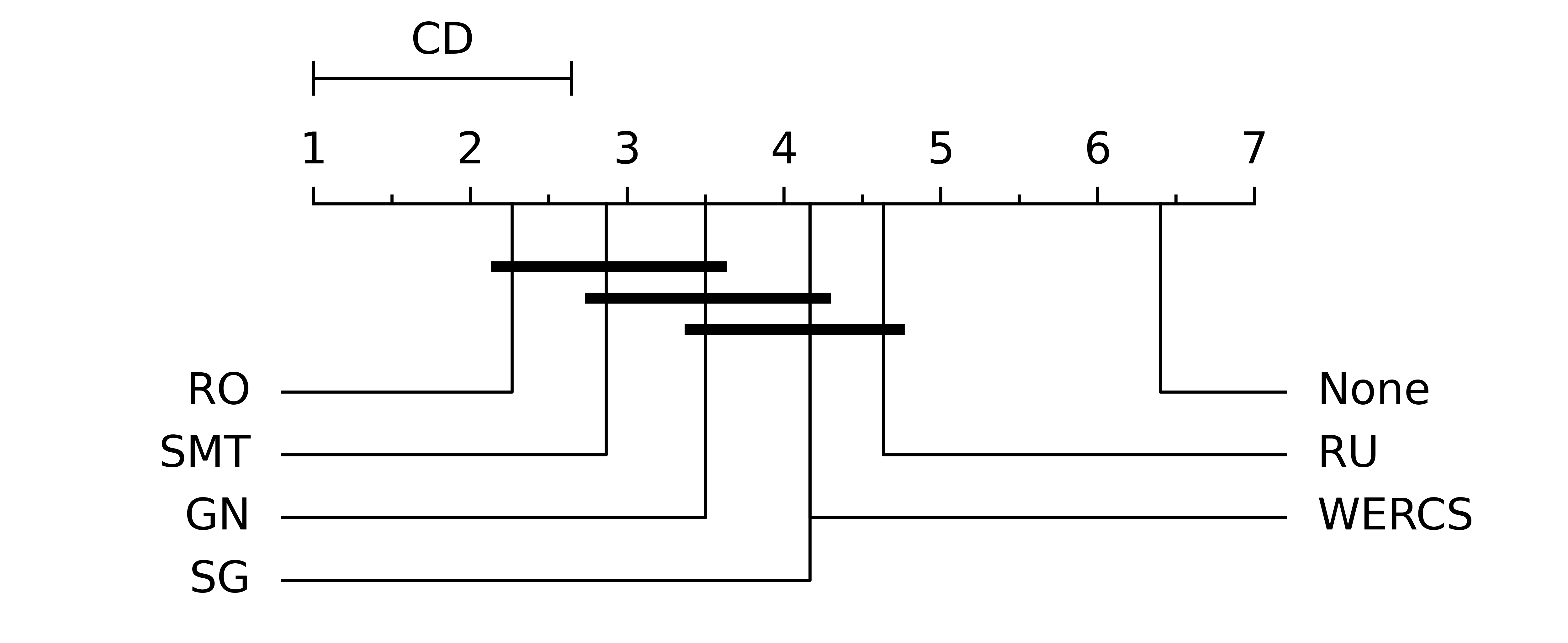}}
\subfigure[Results of the F1-score metric for the XG algorithm\label{CDMLP}]{
\includegraphics[width=6.5cm,height=3.5cm]{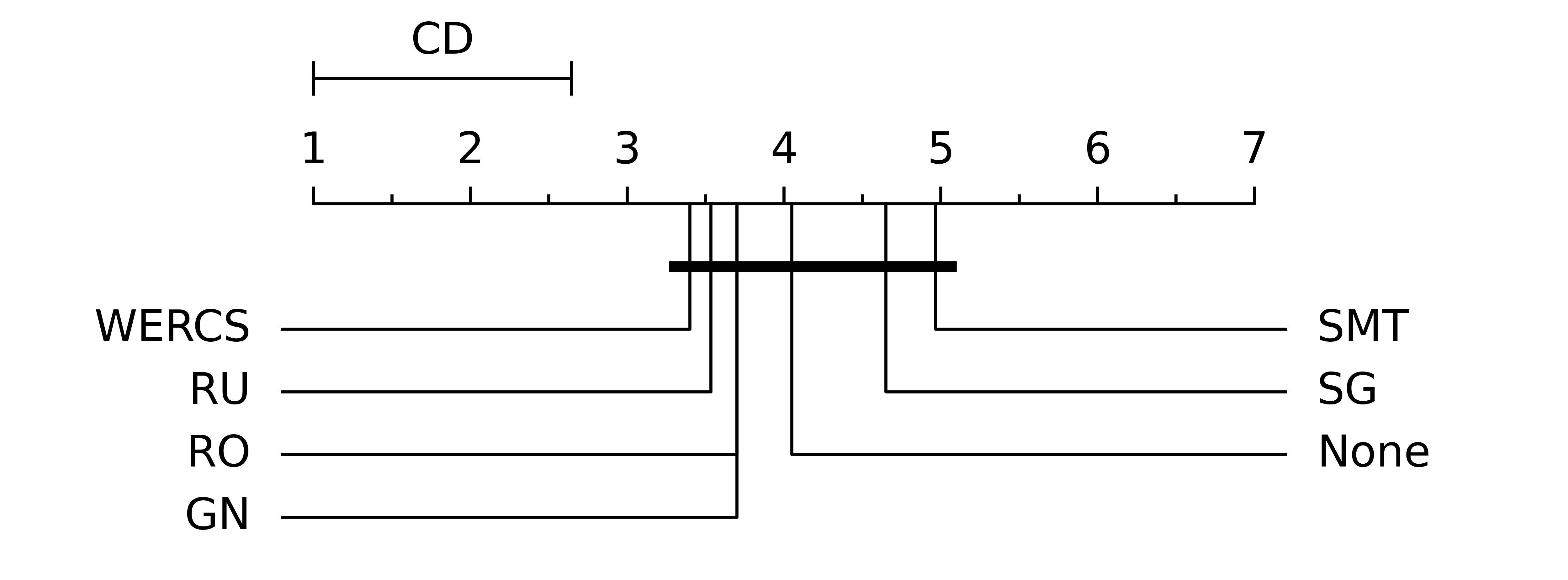}}
\caption{Critical difference diagrams for each learning algorithm considering the F1-score metric}
\label{fig:DC}
\end{figure}

\begin{figure}[h]
\subfigure[Results of the SERA metric for the BG algorithm\label{CDDT}]{
\includegraphics[width=6.5cm,height=3.5cm]{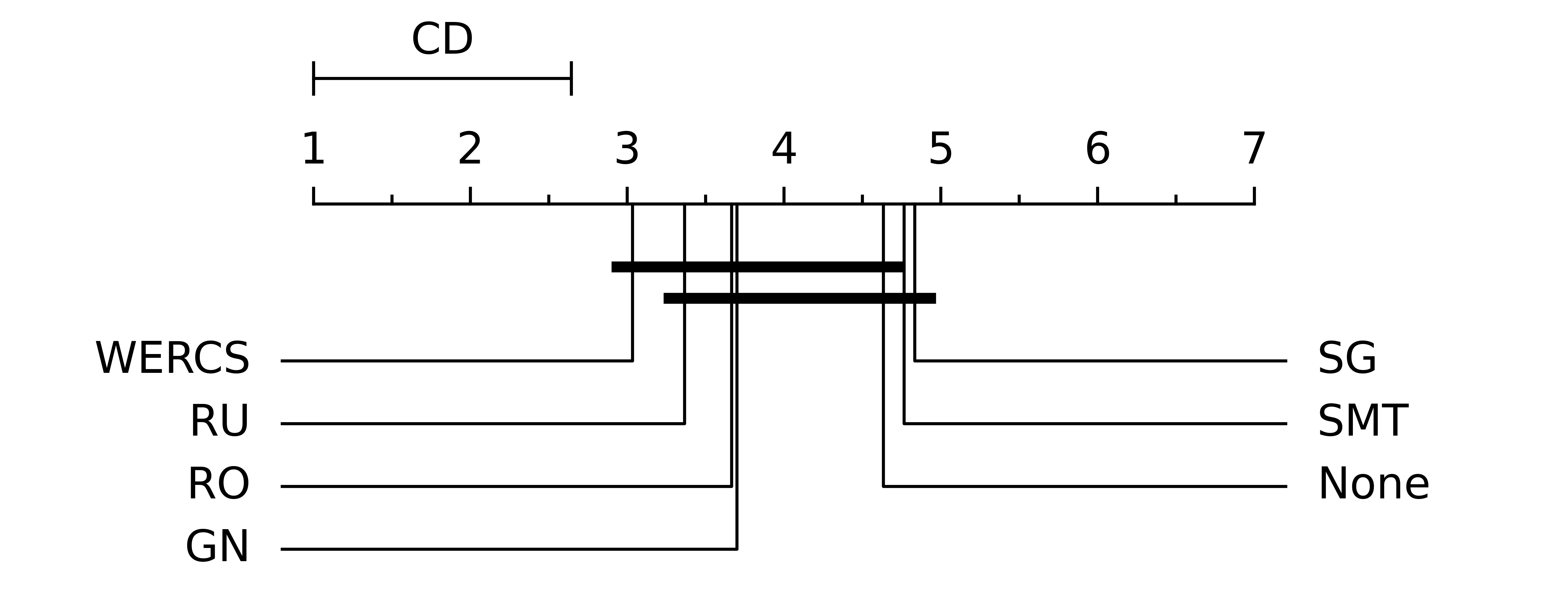}}
\subfigure[Results of the SERA metric for the DT algorithm\label{CDBG}]{
\includegraphics[width=6.5cm,height=3.5cm]{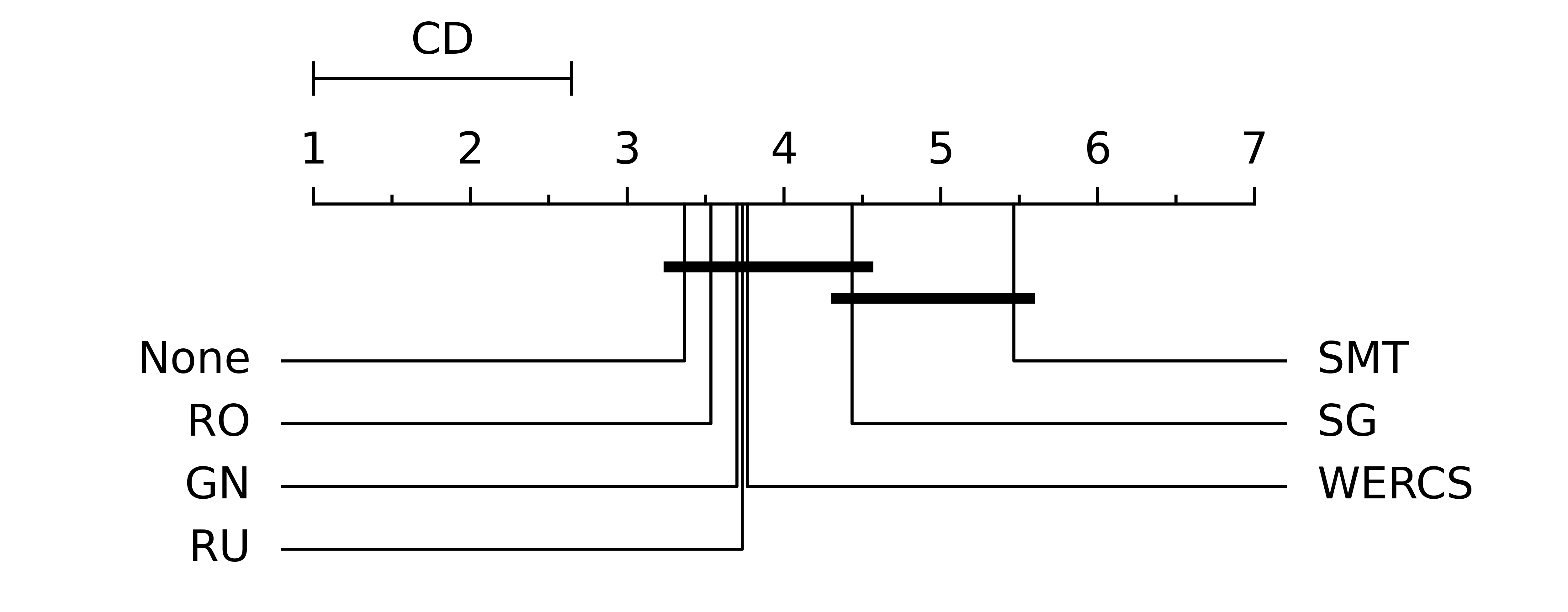}}
\subfigure[Results of the SERA metric for the MLP algorithm\label{CDXG}]{
\includegraphics[width=6.5cm,height=3.5cm]{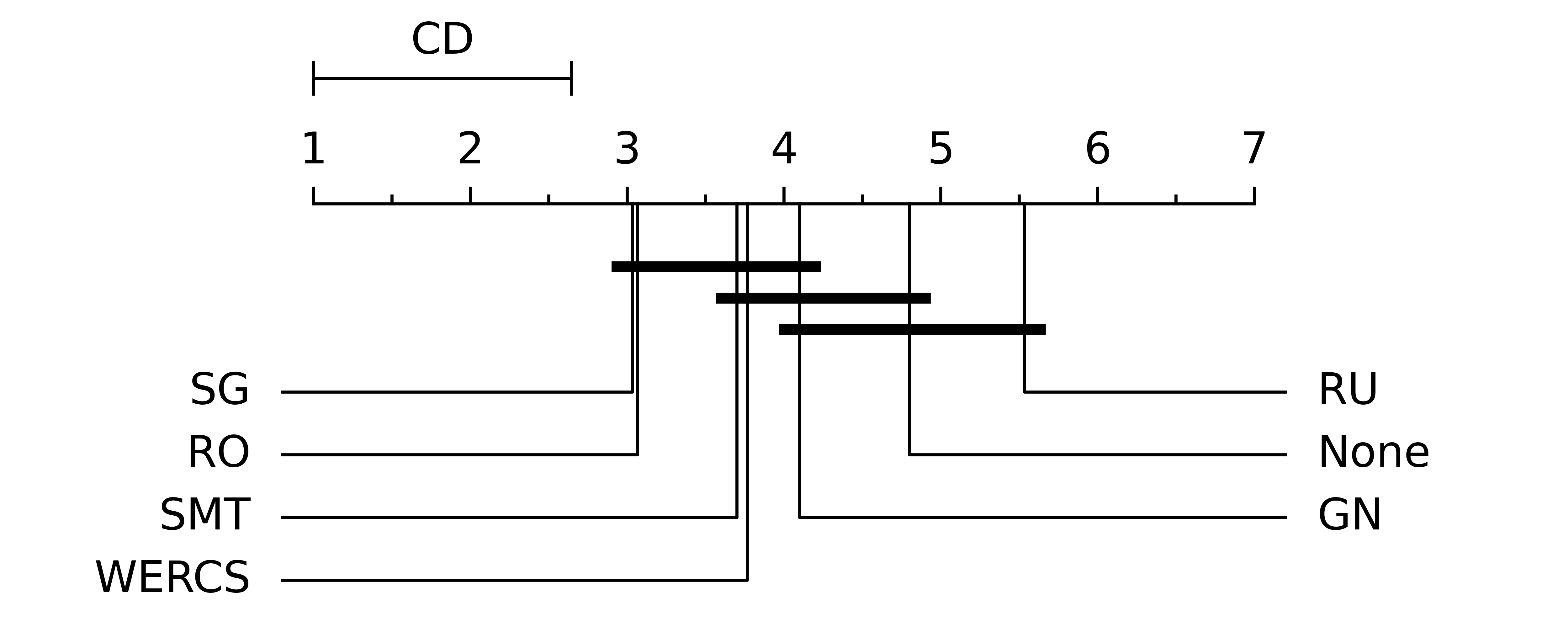}}
\subfigure[Results of the SERA metric for the RF algorithm\label{CDRF}]{
\includegraphics[width=6.5cm,height=3.5cm]{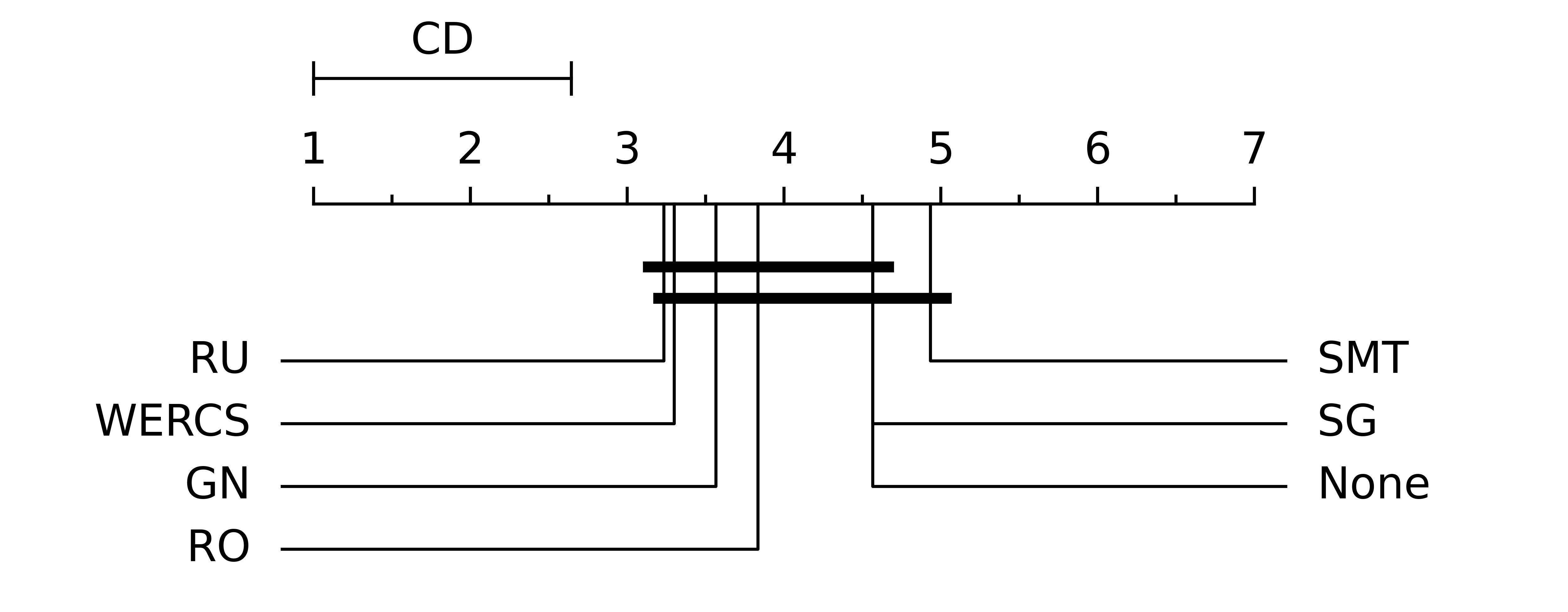}}
\subfigure[Results of the SERA metric for the SVR algorithm\label{CDSVM}]{
\includegraphics[width=6.5cm,height=3.5cm]{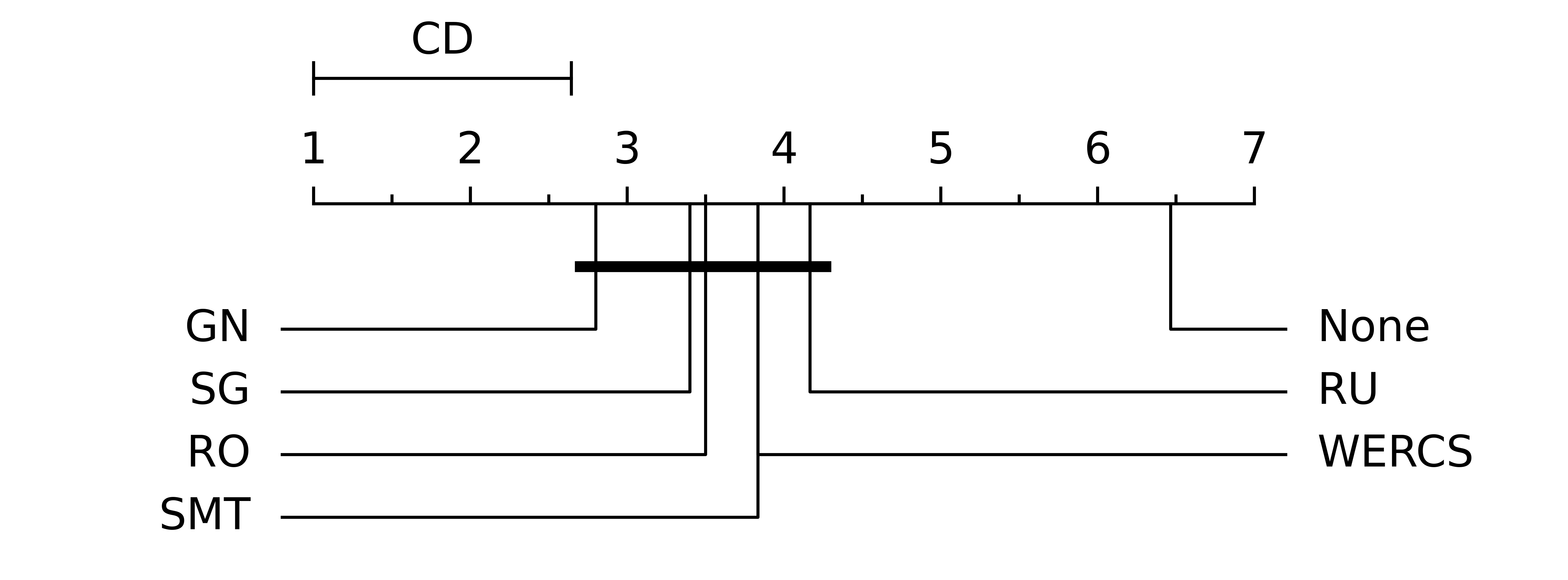}}
\subfigure[Results of the SERA metric for the XG algorithm\label{CDMLP}]{
\includegraphics[width=6.5cm,height=3.5cm]{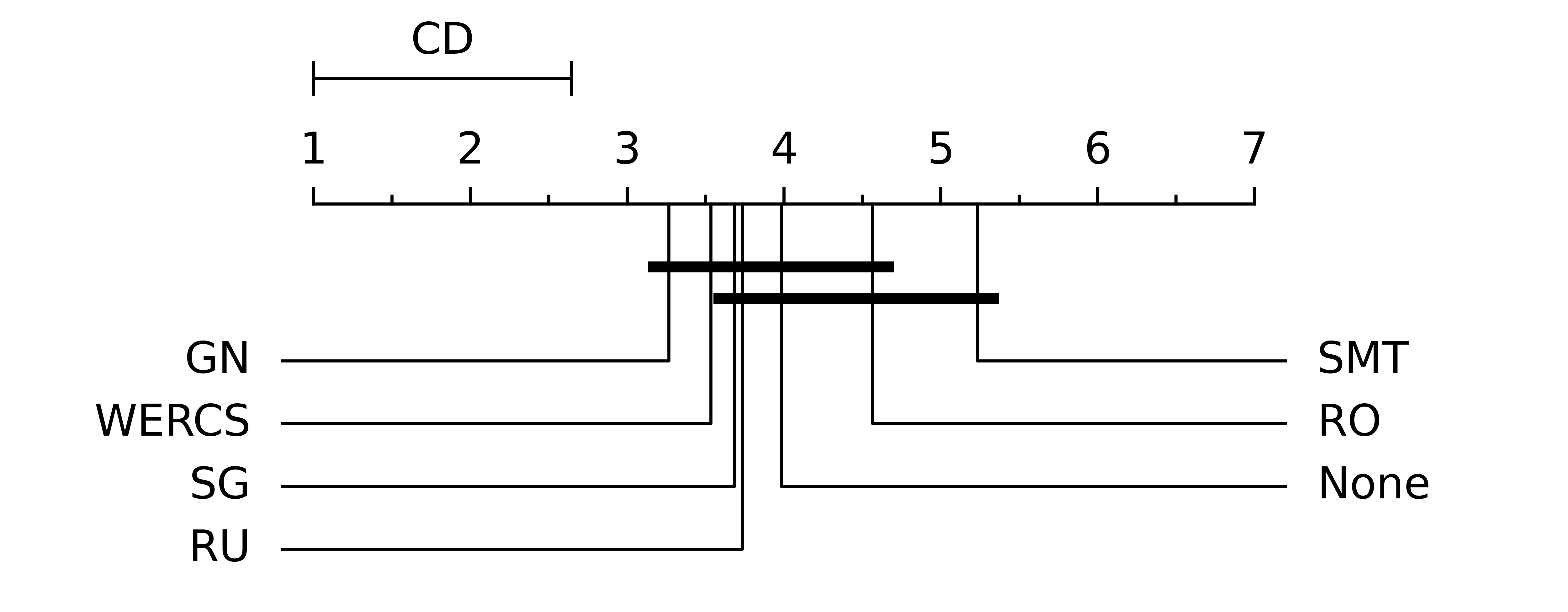}}
\caption{Critical difference diagrams for each learning algorithm considering the SERA metric}
\label{fig:DCSERA}
\end{figure}

To verify which approaches are statistically different, we applied the Nemenyi post-hoc test. Figure~\ref{fig:DC} (a-f) illustrates the critical difference diagrams~\cite{demvsar2006statistical} for each of the learning models, considering the F1-score metric. The horizontal line demonstrates the significance of the difference between the models. Models that are not connected present a significant difference ($p-value <0.05$) in relation to the others. This test once again confirms that, globally, resampling strategies can significantly improve the regressors' performance. The Nemenyi test reveals that the RO obtained the best results and the most significant differences in relation to None (data without any preprocessing) for the metric F1-score. In most cases, the SMT, SG, RU techniques achieve the worst results. Figure~\ref{fig:DCSERA} (a-f) considers the SERA metric; in such a scenario, most of the best results are obtained using the GN strategy, followed by WERCS, given the number of times where the best results were achieved in the critical difference chart.

Another interesting fact is how different learning algorithms perform when no resampling strategy is applied. In both metrics, the DT model achieved better results with the original data sets. Additionally, an interesting aspect involves the ensemble models, Random Forest~(RF) and XGBoost (XG) obtained better results than single models, corroborating the analysis conducted in~\cite{moniz2017evaluation}, which says that ensemble methods provide a better result than single models. Conversely, the SVR and MLP algorithms obtained the worst results, especially when no preprocessing techniques were employed. Thus, it can be concluded that these algorithms are the most affected by having an imbalanced target distribution and require special attention when applied in the imbalanced regression context.

\begin{figure}[H]
    \centering
    \includegraphics[width=16cm]{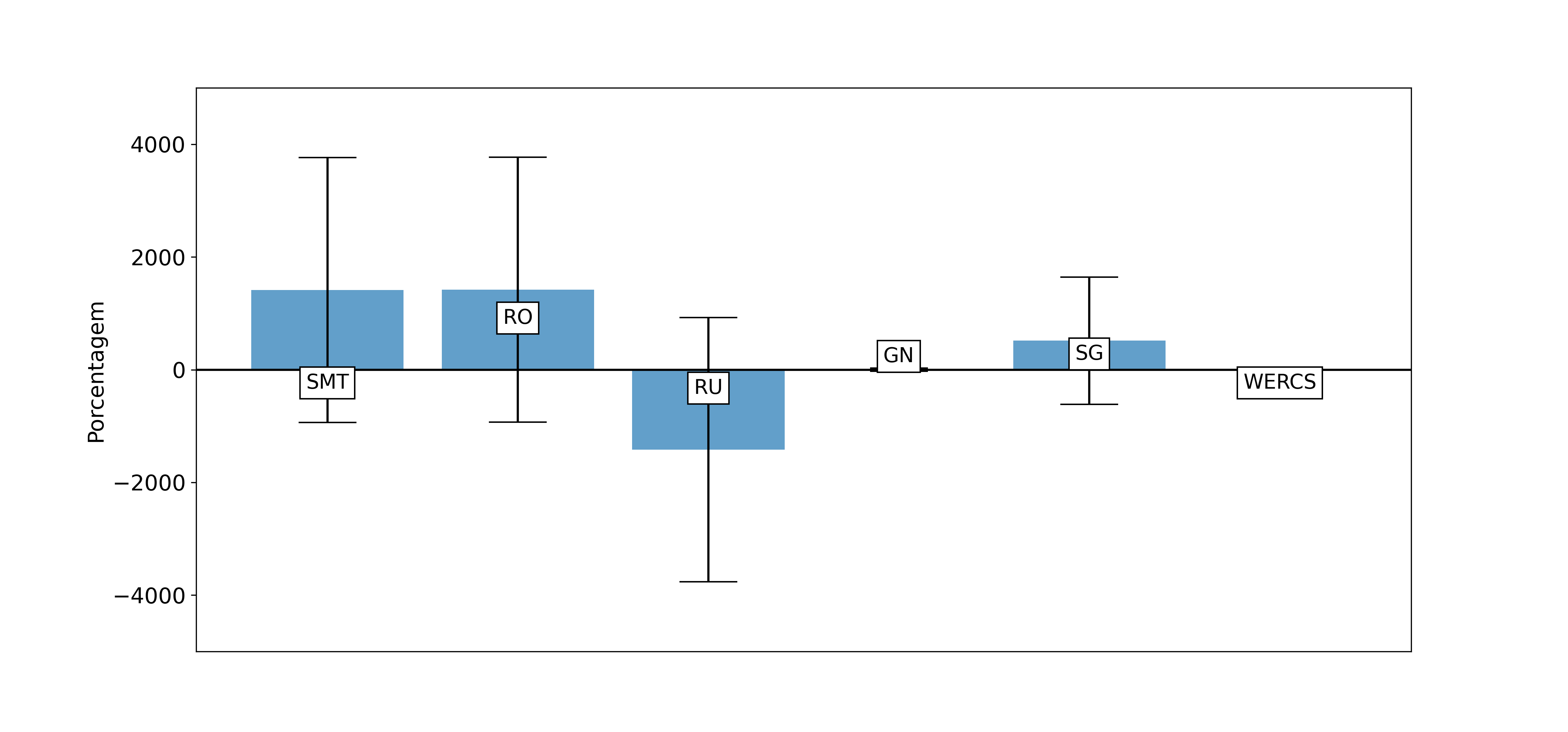}
    \caption{Percentage of increase/decrease in the training set for each resampling strategy.}
    \label{fig:QTD}
\end{figure}

As described in Section~\ref{sec:resampling}, each resampling algorithm uses different heuristics to balance the dataset. Figure~\ref{fig:QTD} illustrates the percentage of increase/decrease in the training examples for each strategy. It was previously concluded that the best results were achieved using the RO, GN, and WERCS strategies. The GN and WERCS strategies present a small percentage of 1.28\% and 2.83\%, respectively. Conversely, the RO presents an increased percentage of 1421.1\%. Therefore, the influence of the number of examples on the results is unclear since the strategies with different percentages of increase/decrease obtained good results. Nonetheless, it may be disadvantageous (from a training time point of view) to use a strategy, such as the RO, that considerably increases the training set. Other strategies also deliver satisfactory results without excessively increasing the training set. More details about the number of instances after the application of the resampling strategies can be found in the supplementary material -- \href{https://github.com/JusciAvelino/imbalancedRegression/blob/main/appendices/Appendix%20C.pdf}{Appendix C}.

\begin{figure}[!h]
    \centering
    \includegraphics[width=15cm]{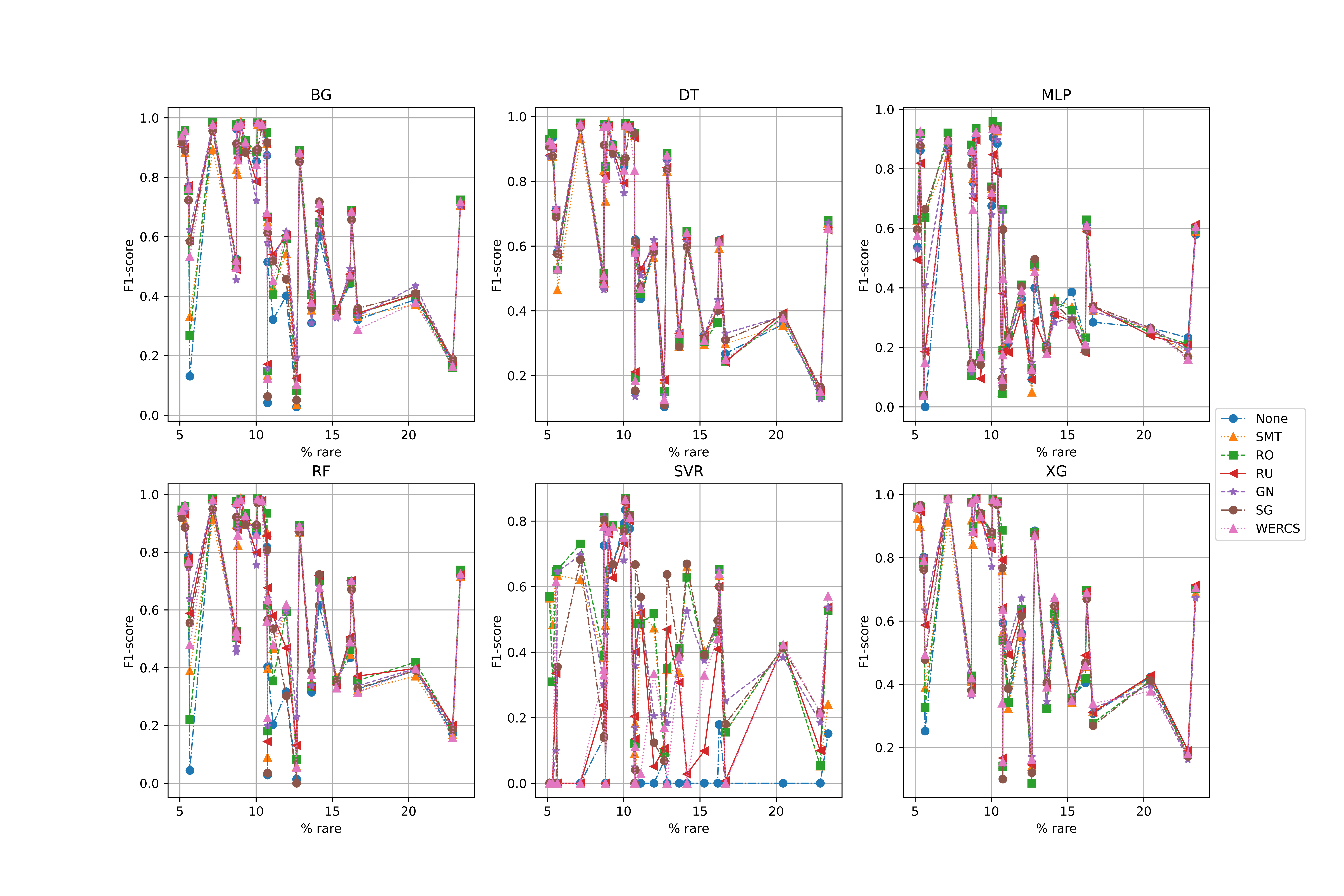}
    \caption{Evolution of the \textit{F1-score} with datasets sorted by percentage of rare cases}
    \label{fig:PRARO}
\end{figure}

\begin{figure}[!h]
    \centering
    \includegraphics[width=15cm]{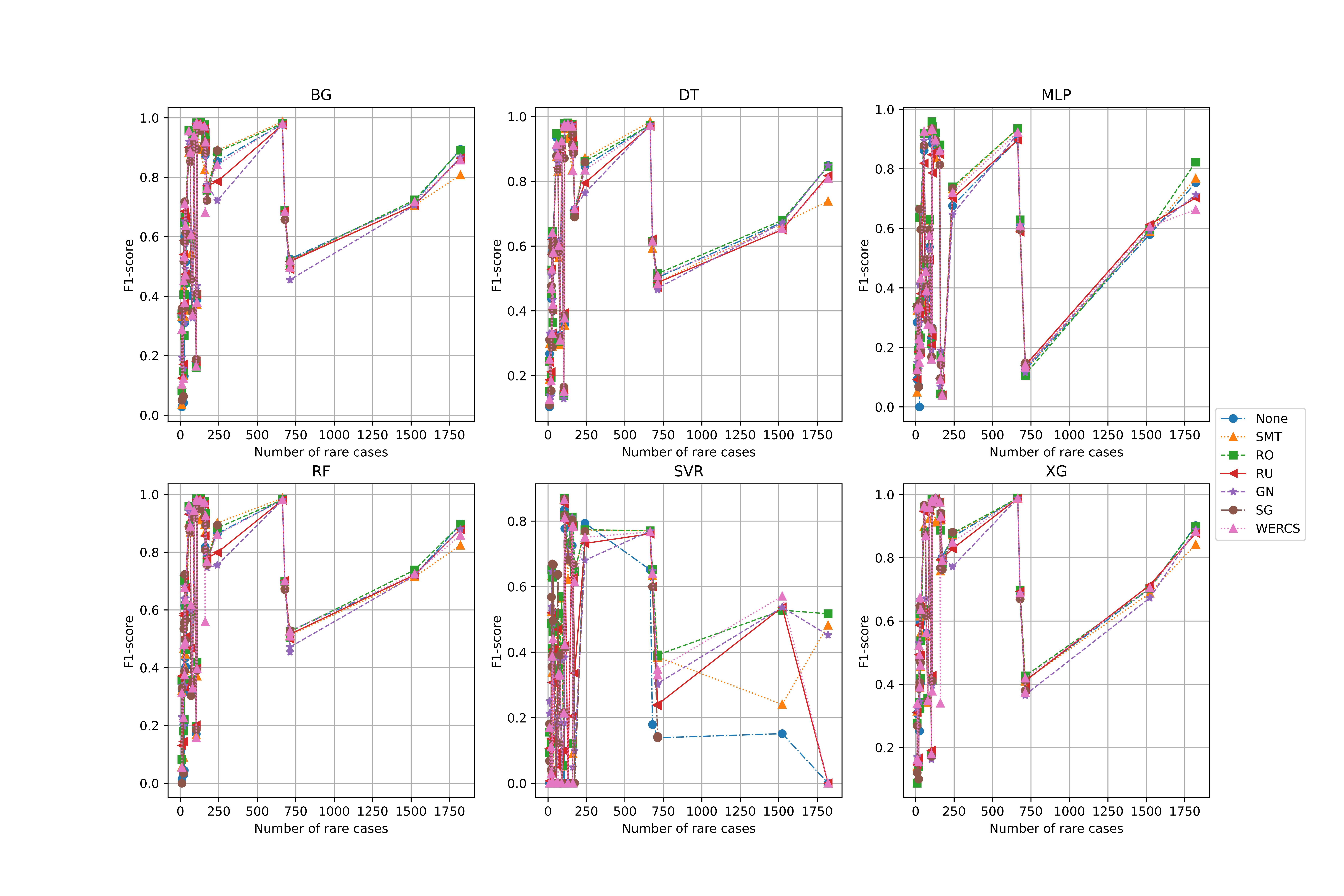}
    \caption{Evolution of the \textit{F1-score} with datasets sorted by number of rare cases}
    \label{fig:NRARO}
\end{figure}

\begin{figure}[!h]
    \centering
    \includegraphics[width=15cm]{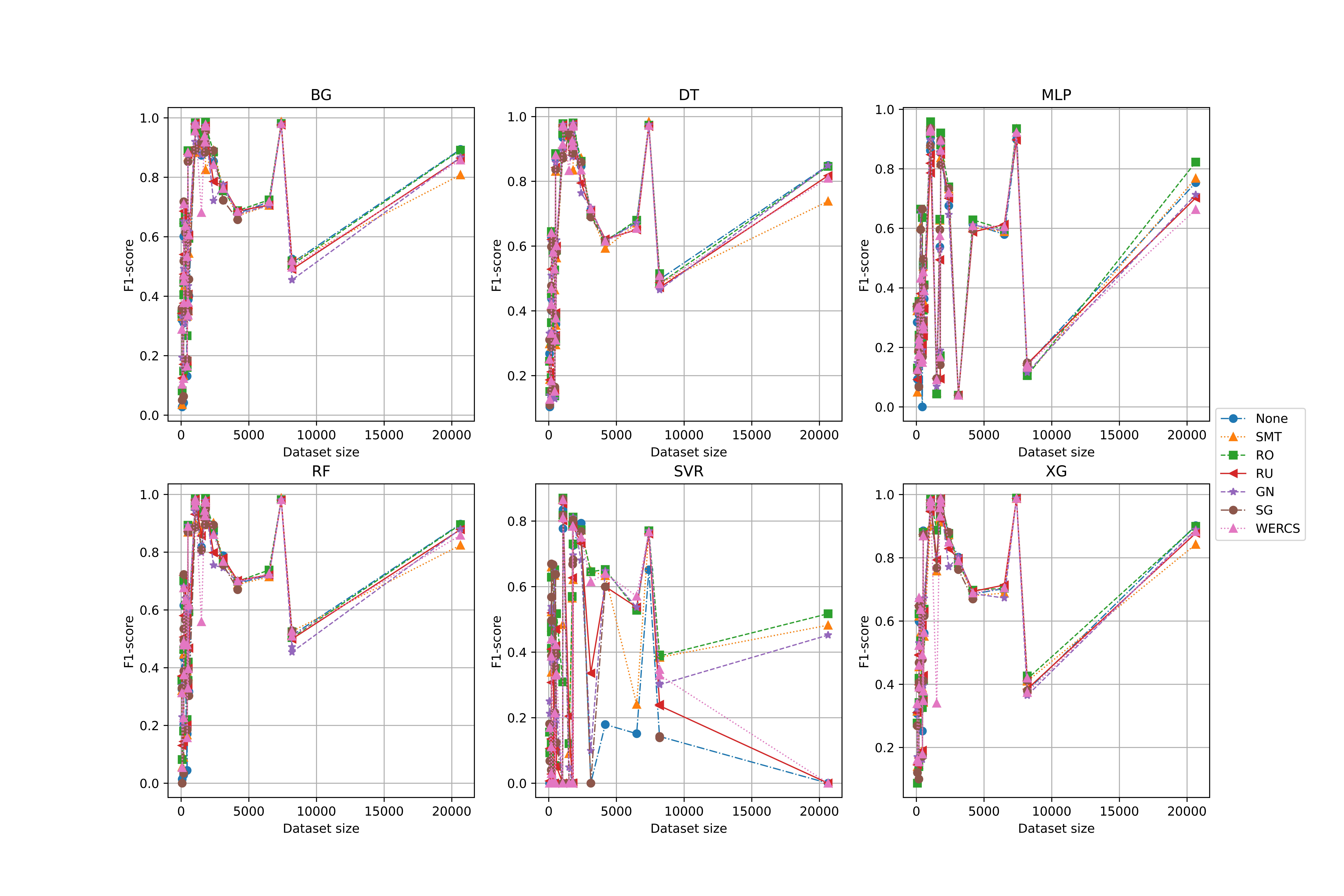}
    \caption{Evolution of the \textit{F1-score} with datasets sorted by size}
    \label{fig:size}
\end{figure}

\begin{figure}[!h]
    \centering
    \includegraphics[width=15cm]{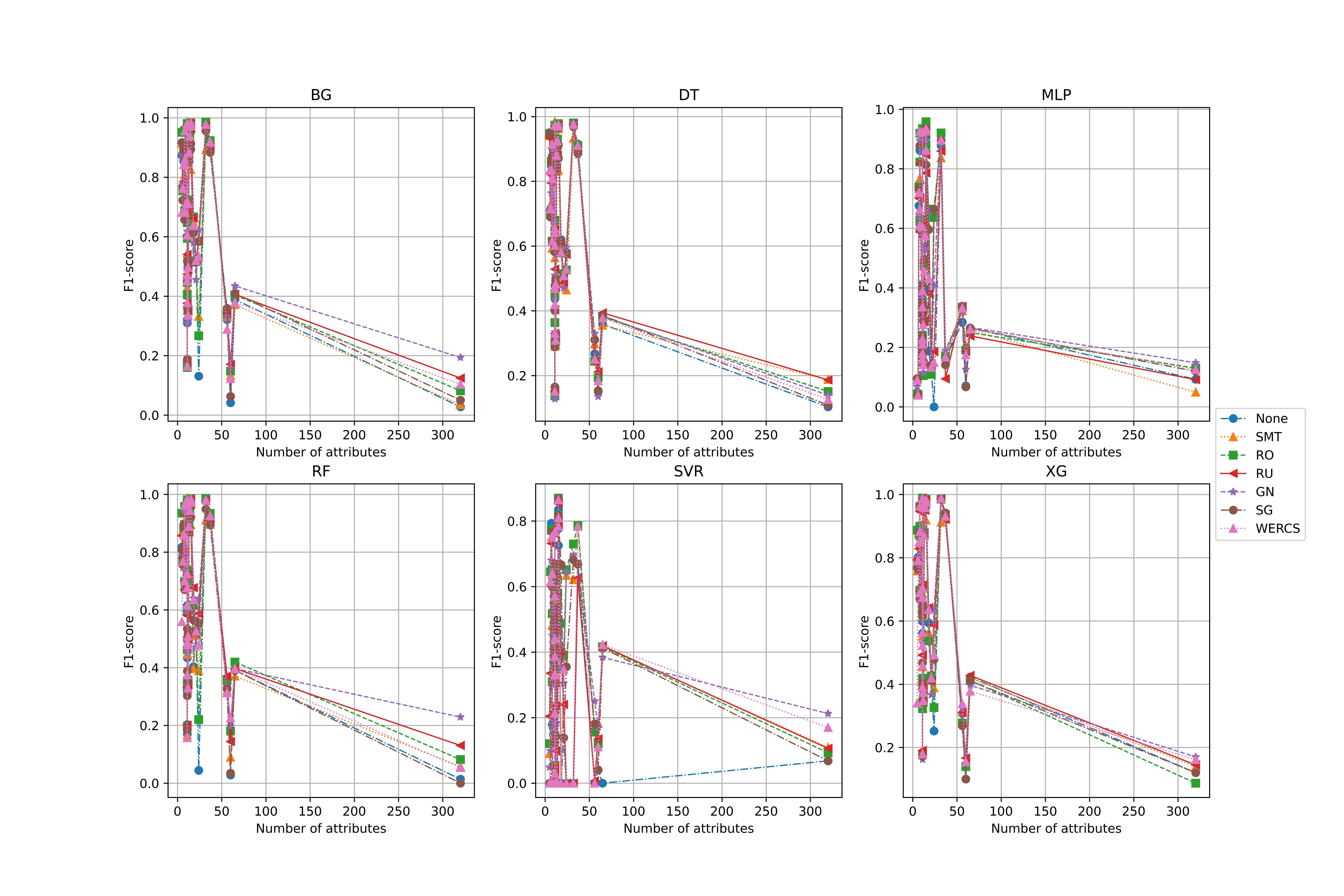}
    \caption{Evolution of the \textit{F1-score} with datasets sorted by number of attributes}
    \label{fig:nattrib}
\end{figure}

\begin{figure}[!h]
    \centering
    \includegraphics[width=15cm]{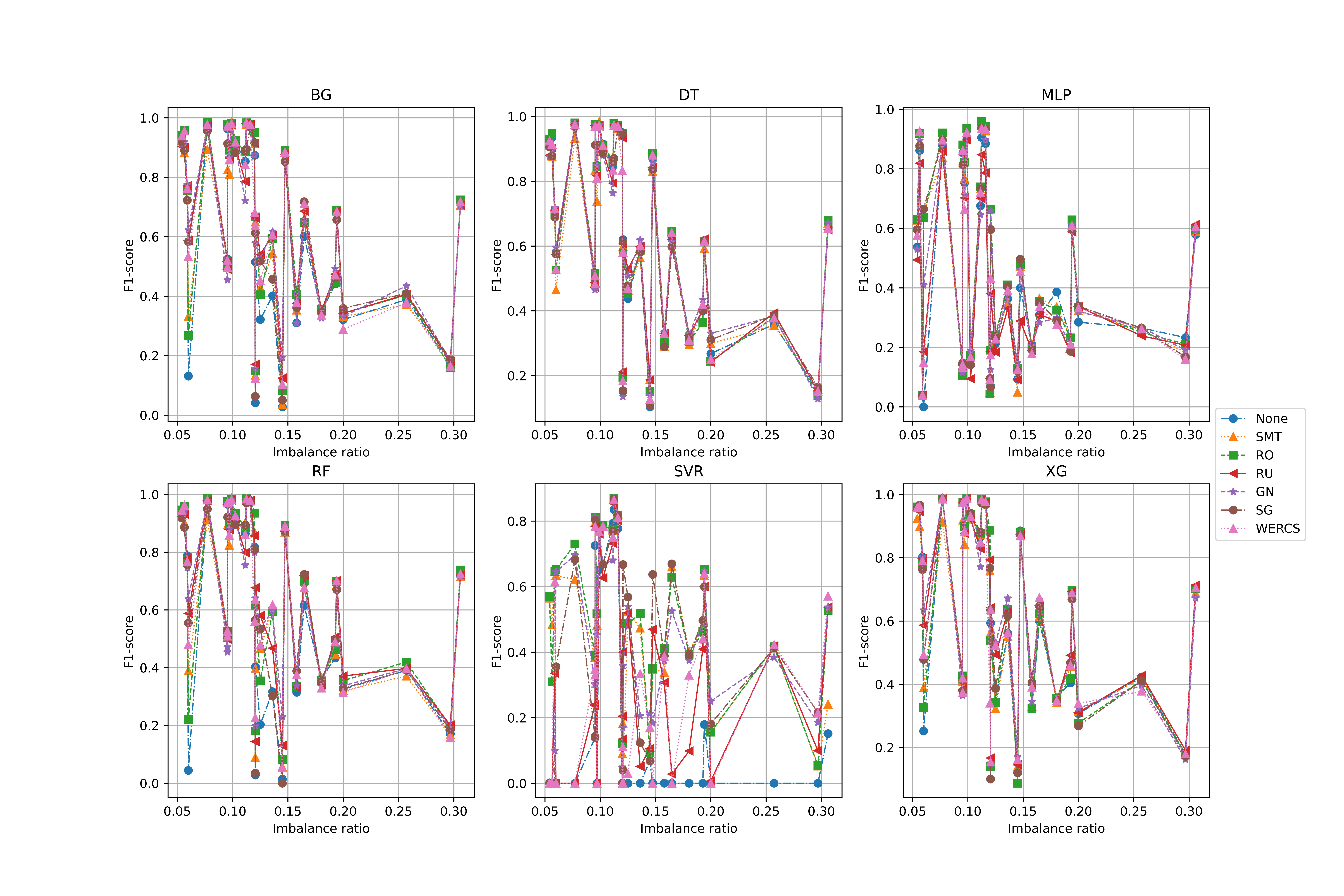}
    \caption{Evolution of the \textit{F1-score} with datasets sorted by imbalance ratio}
    \label{fig:ratio}
\end{figure}

Figures~\ref{fig:PRARO},~\ref{fig:NRARO},~\ref{fig:size},~\ref{fig:nattrib} and~\ref{fig:ratio} present the F1-score results arranged according to some dataset characteristics in a bid to assess their impact on the performance of the models. The following characteristics were assessed: percentage of rare cases, number of rare cases, dataset size, number of attributes, and imbalance ratio. The imbalance ratio is calculated as the ratio between the number of rare cases ($D_R$) and the number of normal cases ($D_N$), i.e., $\frac{|D_R|}{|D_N|}$. Each box represents a regression model (BG, DT, MLP, RF, SVR and XG), and each point represents a specific set of data, and each line represents a resampling strategy (None, SMT, RO, RU, GN, SG and WERCS).

The results presented in Figure \ref{fig:PRARO} correspond to the same ordering provided in Table \ref{tab:ds}, where the datasets are arranged in decreasing order of the percentage of rare cases. In such conditions, it is not possible to find any pattern. Therefore, it is unclear how this aspect relates to the model's performance. Figures~\ref{fig:NRARO} and~\ref{fig:size} are arranged according to the number of rare cases and the dataset size, respectively. These circumstances reveal that the smaller datasets with a lower number of rare cases represent the hardest tasks, as observed in~\cite{branco2019pre}. Figure~\ref{fig:nattrib} illustrates the evolution of F1 considering the number of attributes in each dataset. In some instances, it is noticeable that datasets with fewer features exhibit superior performance. Finally, in Figure~\ref{fig:ratio}, the datasets are sorted according to their respective imbalance ratios. The regression models with all resampling strategies face more significant challenges when dealing with datasets exhibiting higher imbalance ratios. This difficulty arises because higher imbalance ratios mean the rare cases are significantly underrepresented compared to the normal cases. As a result, the model may struggle to learn the underlying patterns and become biased toward the normal cases. 


For all the evaluated dataset characteristics, the behavior of the resampling strategies is quite similar, resulting in an overlap of the graph's line. For better clarity, another analysis is conducted considering the best F1-score for each dataset. The figures in \href{https://github.com/JusciAvelino/imbalancedRegression/blob/main/appendices/Appendix%20D.pdf}{Appendix D} present the best F1-score for each dataset, considering the dataset characteristics. With this, we can visualize how the data characteristics affect the performance of the top models. The percentage of rare cases does not exhibit a clear pattern. Thus, concluding whether this characteristic affects the model's performance is challenging. Regarding the number of rare cases and the dataset size, models achieve better performance when there are more rare cases and a larger dataset. When we consider the number of attributes, we observe that a higher number leads to better model performance. As for the imbalance ratio, the higher the imbalance ratio, the worse the model's performance.

\section{Lessons Learned}
\label{sec:lessons}

Different approaches have been proposed in a bid to solve the imbalanced problem in the context of regression, including resampling strategies. Our research introduced a review and an experimental study of the main resampling strategies for dealing with imbalanced regression problems. In this section, the research questions are revisited and answered succinctly.

\begin{enumerate}
  \item \textbf{Is it worth using resampling strategies?}
    
    \vspace{0.2cm}
    
    We answer this question by accounting for the number of times that each strategy won (Tables~\ref{tab:venceuf1} and~\ref{tab:venceusera}). For both metrics, four of the resampling strategies used won more times than when no resampling strategy was used. Furthermore, the Nemenyi post-hoc statistical tests performed (Figures~\ref{fig:DC} and~\ref{fig:DCSERA}) demonstrate that many resampling strategies are statistically better as compared to the absence of a strategy. Therefore, it is advantageous to use (some) resampling strategies.
    
    \vspace{0.2cm}
    
  \item \textbf{Which resampling strategies influence the predictive performance the most?}
    
    \vspace{0.2cm}
    
    Considering the F1-score metric, the RO and GN strategies positively influenced the results of the learning algorithms. As for the SERA metric, the GN and WERCS techniques are the best strategies. Statistically, in general, the GN, RO, and WERCS strategies held the best results (Figures~\ref{fig:DC} and~\ref{fig:DCSERA}). Conversely, in terms of predictive performance, the SMT, SG, RU techniques achieve the worst results..
    
    \vspace{0.2cm}

  \item \textbf{Does the choice of best strategy depend on the dataset, the learning model, and the metrics used?}
    
    \vspace{0.2cm}
  
   Most of the datasets used have distinct preferences regarding the combination of the best regression model and resampling strategy (Tables~\ref{tab:mpf1} and~\ref{tab:mpsera}). For the regression models, different resampling strategies can reach better results. As for the metrics, both agree that the GN is a good resampling strategy. Nonetheless, there are cases of disagreement between them.


    \vspace{0.2cm}
      
  \item \textbf{Does the number of training examples resulting from each strategy influence the results?}
  
    \vspace{0.2cm}
    
    Given that the best results were obtained using the GN, RO, and WERCS strategies, which have different percentages (1.28\%,  1421.1\%, 2.83\%, respectively) of increase/decrease in the training examples (Figure~\ref{fig:QTD}), the influence of the number of examples on the results is not clear. Nonetheless, it may not be advantageous (from a training time point of view) to use a strategy like the RO, which considerably increases the training set, as other strategies also deliver equivalent results without this excessive increase.
  
    \vspace{0.2cm}
    
  \item \textbf{Do the features of the data (percentage of rare cases, number of rare cases, dataset size, number of attribues and imbalance ratio) impact the predictive performance of the models?}
    
    \vspace{0.2cm}

    In the studies performed, the percentage of rare cases did not have a clear impact on the results. On the other hand, considering the dataset size and number of rare cases, it could be seen that the smaller datasets with fewer rare cases correspond to the most difficult tasks.  Models demonstrate superior performance in datasets with fewer features. Lastly, concerning the imbalance ratio, regression models encounter more significant challenges with a higher imbalance ratio. The results for this question are shown in Figures \ref{fig:PRARO}, \ref{fig:NRARO}, \ref{fig:size}, \ref{fig:nattrib} and \ref{fig:ratio}. \href{https://github.com/JusciAvelino/imbalancedRegression/blob/main/appendices/Appendix%20D.pdf}{Appendix D} presents the evolution of the best F1-score for each dataset characteristic, providing a clearer view of the impact of these dataset characteristics on model performance.

\end{enumerate}

\section{Conclusion}
\label{sec:conclusion}

This work reviews and performs a comparative study of data resampling strategies for handling imbalanced regression problems. We reviewed six state-of-the-art resampling strategies for regression based on three approaches: i) under-sampling, ii) oversampling, and iii) a mix  of undersampling and  oversampling, while discussing  the advantages and drawbacks of each existing technique.

Then, we performed an extensive experimental analysis comprised of  6 regression algorithms and 7 scenarios (6 resampling strategies and not using resampling)  that can guide the development of new strategies to solve the imbalanced regression problem. Our experimental results demonstrate that it is important to use a resampling technique for most models as resampling techniques lead to statistically better results. The experimental study also shows that no resampling technique outperforms all others. Furthermore, choosing the best resampling technique depends on three main factors: the learning algorithm, the dataset, and the performance metric used to assess the model's performance. 



Further studies should address the recommendation of combining resampling strategies with a regression model for each specific dataset. Another element worth addressing is the dataset characteristics, which should be investigated through data complexity measures~\cite{lorena2018data} in order to assess the adverse effects of these features on prediction performance. Moreover, an essential point to address involves proposing a new relevance function since currently, only one definition exists. This proposal aims to conduct studies and comparisons regarding the definition of an imbalanced regression dataset.

\backmatter





\bmhead{Acknowledgments}

This work was partially supported by several Brazilian agencies: ({\it Fundação de Amparo à Ciência e Tecnologia do Estado de Pernambuco} (FACEPE) and {\it Conselho Nacional de Desenvolvimento Científico e Tecnológico} (CNPq)), and the \'{E}cole de technologie sup\'{e}rieure (\'{E}TS Montr\'{e}al). 

\section*{Declarations}

\subsection*{Conflict of interest}

The authors have no competing interests to declare that are relevant to the content of this article.

\bibliography{sn-bibliography}

\clearpage

\end{document}